\pdfoutput=1

\documentclass[11pt]{article}

\usepackage[]{EMNLP2023}

\usepackage{times}
\usepackage{latexsym}

\usepackage[T1]{fontenc}

\usepackage[utf8]{inputenc}

\usepackage{microtype}

\usepackage{inconsolata}
\usepackage{listings}
\usepackage[utf8]{inputenc} 
\usepackage[T1]{fontenc}    
\usepackage{hyperref}       
\usepackage{url}            
\usepackage{booktabs}       
\usepackage{amsfonts}       
\usepackage{nicefrac}       
\usepackage{microtype}      
\usepackage{xcolor}         
\usepackage{xspace}
\usepackage{multirow}
\usepackage{makecell}
\usepackage{colortbl}
\usepackage{graphicx}
\usepackage{comment}

\newcommand{\model}{\textsc{NovaCOMET}\xspace}
\newcommand{\modelplaus}{\textsc{NovaCOMET}$_{\textit{critic}}$\xspace}

\newcommand{\data}{\textsc{NovAtomic}\xspace}
\newcommand{\dataset}{\textsc{NovAtomic}\xspace}

\newcommand{\atomictenx}{\textsc{ATOMIC10x}\xspace}
\newcommand{\atomictwenty}{\textsc{Atomic}$^{2020}$\xspace}

\newcommand{\modelfilter}[1]{\textsc{NovaCOMET}$_{filter-#1}$\xspace}
\newcommand{\modelcrit}{\textsc{NovaCOMET}$_{crit}$\xspace}
\newcommand{\modelgen}{\textsc{NovaCOMET}$_{base}$\xspace}

\newcommand{\modelquarkg}[1]{\textsc{NovaCOMET}$_{\textrm{rc} #1}$\xspace}

\definecolor{headcolor}{HTML}{018161}
\definecolor{relationcolor}{HTML}{d95f02}
\definecolor{tailcolor}{HTML}{6560a3}

\title{NovaCOMET: Open Commonsense Foundation Models with Symbolic Knowledge Distillation}

\author{Peter West\textsuperscript{$\dagger$$\ddagger$} \hspace{.1cm} \textbf{Ronan Le Bras}\textsuperscript{$\ddagger$} \hspace{.1cm} \textbf{Taylor Sorensen}\textsuperscript{$\dagger$$\ddagger$} \hspace{.1cm} \\ \textbf{Bill Yuchen Lin}\textsuperscript{$\ddagger$} \hspace{.1cm} \textbf{Liwei Jiang}\textsuperscript{$\dagger$$\ddagger$}  \hspace{.1cm} \textbf{Ximing Lu}\textsuperscript{$\dagger$$\ddagger$} \hspace{.1cm} \textbf{Khyathi Chandu}\textsuperscript{$\ddagger$} \hspace{.1cm} \\ \textbf{Jack Hessel}\textsuperscript{$\ddagger$} \hspace{.1cm} \textbf{Ashutosh Baheti}\textsuperscript{$\ddagger$} \hspace{.1cm} \textbf{Chandra Bhagavatula}\textsuperscript{$\ddagger$} \hspace{.1cm} 
\textbf{Yejin Choi \textsuperscript{$\dagger$$\ddagger$}}\\
  \textsuperscript{$\dagger$}Paul G. Allen School of Computer Science \& Engineering, University of Washington\\
  \textsuperscript{$\ddagger$}Allen Institute for Artificial Intelligence\\
  \texttt{\href{mailto:pawest@cs.washington.edu}{pawest@cs.washington.edu}}}

\begin{document}
\maketitle
\begin{abstract}
We present \model, an open commonsense knowledge model, that combines the best aspects of knowledge models and general task models. Compared to previous knowledge models, \model allows open-format relations enabling direct application to reasoning tasks; compared to general task models like Flan-T5, \model explicitly centers knowledge, enabling superior performance for commonsense reasoning. 

\model leverages the knowledge of opaque proprietary models to create an open knowledge pipeline. First, knowledge is symbolically distilled into \dataset, a publicly-released\footnote{Our resources are available at \url{novacomet.dev}} discrete knowledge graph which can be audited, critiqued, and filtered.  Next, we train \model on \dataset by fine-tuning an open-source pretrained model.  \model uses an open-format training objective, replacing the fixed relation sets of past knowledge models, enabling  
arbitrary structures within the data to serve as inputs or outputs. 

The resulting generation model, optionally augmented with human annotation, matches or exceeds comparable open task models like Flan-T5 on a range of commonsense generation tasks. \model serves as a counterexample to the contemporary focus on instruction tuning only, demonstrating a distinct advantage to explicitly modeling commonsense knowledge as well. 
 
\end{abstract}

\section{Introduction}
\label{sec:intro}

\begin{figure}
    \centering
    \includegraphics[width=0.99\linewidth]{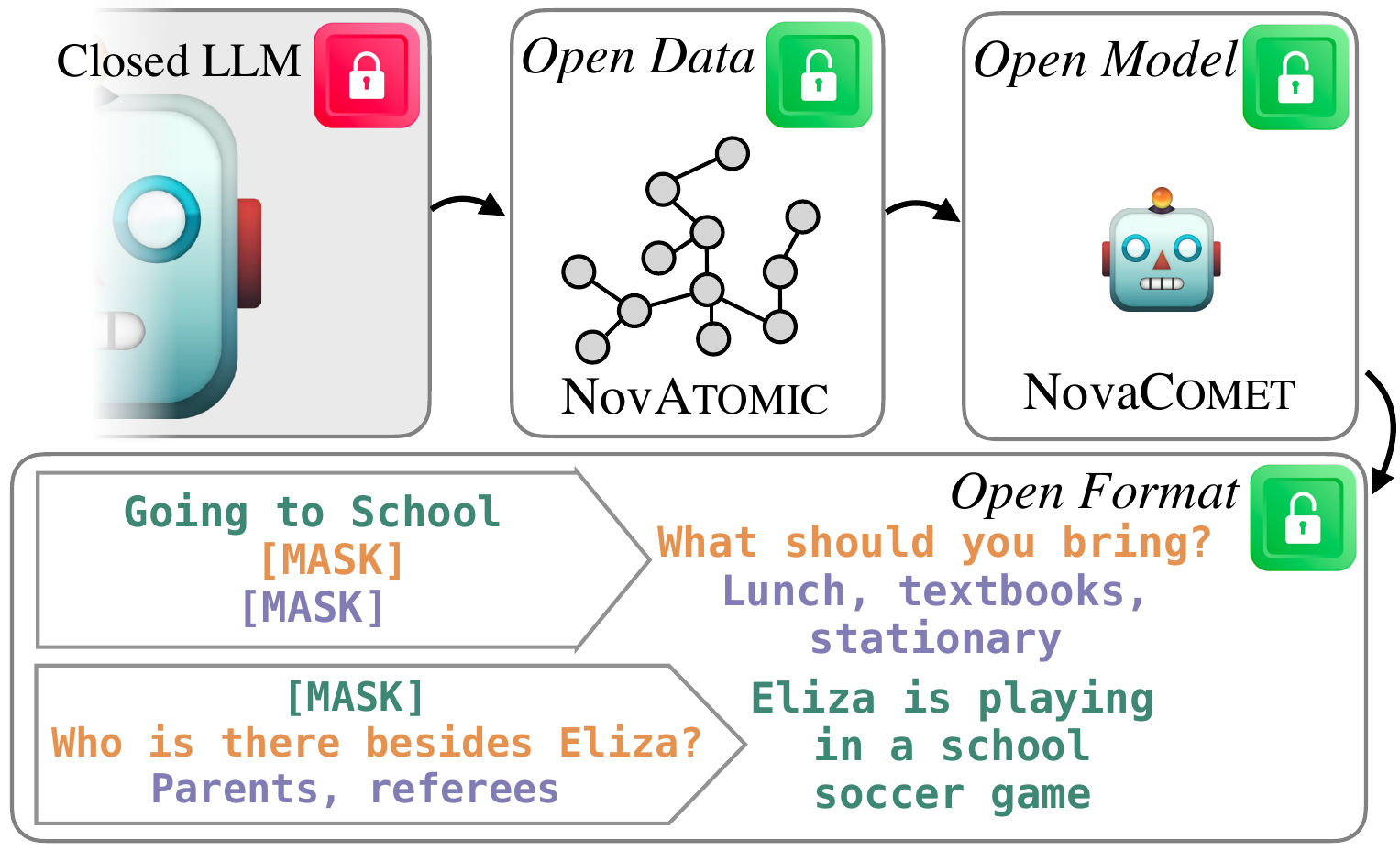}
    \caption{We leverage opaque-but-powerful proprietary LLMs into an open commonsense pipeline by: (i) creating an auditable knowledge base \dataset that
    gives fine-grained control over included knowledge, 
   (ii) ensuring the generated knowledge uses a higher-coverage open-format with natural-language queries as relations and flexible mask-filling to allow for more open commonsense use-cases, 
   (iii) demonstrating the effectiveness of (i) and (ii) via \model's superior performance on a number of tasks, under both automatic and human evaluations.
    }
    \label{fig:1}
\end{figure}




We present \model, an open commonsense knowledge model combining the advantages of both knowledge models and general task models. 
\model models commonsense knowledge with an open format, allowing it to be applied to general reasoning tasks in contrast to previous knowledge models. Compared to simply training models to be open task solvers (e.g. instruction tuning) we find that explicitly modeling knowledge in \model also provides a distinct advantage, with \model showing similar or superior performance to comparable open task models on a range of commonsense reasoning benchmarks.

For \model, we leverage opaque, proprietary models like ChatGPT or GPT-4~\cite{Ouyang2022TrainingLM,OpenAI2023GPT4TR} as the knowledge source in an open commonsense pipeline (Figure~\ref{fig:1}). Such models have demonstrated remarkable commonsense ability~\cite{bubeck2023sparks,Bian2023ChatGPTIA} yet, closed and opaque, their direct usefulness for studying commonsense is limited. Without information about training or direct access to the model, it is impossible to study \emph{where} 
reported gains come from---e.g. the extent of test set contamination with benchmarks. 

In our work, we use these models first to generate an \textbf{open} knowledge base (\textbf{\dataset}, \S\ref{subsec:data}), which can be analyzed, improved, and verified against test set contamination.
Next, we train an \textbf{open} commonsense model (\textbf{\model}, \S\ref{subsec:training}) on this knowledge: the underlying data and code will be released along with the model for the study of commonsense. This allows future testing of \model (and of other models based on \dataset) to analyze the training set—essentially allowing us to distill information from a base LLM into an auditable format.

In training \model, we also use an \textbf{open} format: compared to previous knowledge models which use a fixed relation set and training order (head + relation$\rightarrow$ tail) we use natural language queries as relations, and allow masked generation of all aspects of the data. 
This allows our model to be used in a wide range of general reasoning tasks, thus addressing a significant limitation of prior knowledge models that are limited to downstream applications capable of effectively leveraging their restricted set of relations.
Enabling an open format also allows the knowledge generation to focus on pertinent aspects of the context, rather than forcing the generation of inferences for arbitrary, potentially irrelevant relations.

Following past work on symbolic knowledge distillation \cite{west-etal-2022-symbolic}, we also use \dataset as the basis for training a plausibility model with human annotations (\S\ref{subsec:plaus_data}), and study how this can improve \model 
(\S\ref{subsec:training}). 

We test \model on a range of commonsense generation tasks, and find that it consistently outperforms general task models of comparable size, such as Flan-T5$_{xxl}$ \cite{Chung2022flan} and T0 on commonsense tasks like abductive infilling and explanation generation. Furthermore, we assess the ability of our plausibility model to handle general commonsense QA tasks and observe that it achieves comparable or superior discriminative performance on a range of tasks. \model will serve as an open resource for studying commonsense, and an example of the advantage of explicitly modeling commonsense knowledge in contrast to general task modeling alone.

\section{\model: open commonsense models}

\model is a large-scale, open commonsense model that can handle both explicit knowledge generation, and tasks that require commonsense reasoning. 

\model is trained with symbolic knowledge distillation \cite{West2021SymbolicKD} by combining the commonsense data generated by large language models (\S\ref{subsec:data}) with high-quality annotations of knowledge plausibility (\S\ref{subsec:plaus_data}). We experiment with multiple methods for combining generated data with plausibility information (indicating how likely a given knowledge is) to train the final model, \model (\S\ref{subsec:training}).

\subsection{Generating Open Commonsense Data}
\label{subsec:data}

Following symbolic knowledge distillation \cite{West2021SymbolicKD}, we distill large quantities of high-quality knowledge from very large, general foundation models (\S\ref{subsubsec:data_gen}) -- we call the resulting dataset \dataset. One major difference from previous knowledge graphs is that we allow an open relation set, in the form of queries rather than fixed relation tokens. While commonsense knowledge often takes a \emph{head, relation, tail} format with a fixed set of discrete relations (e.g. \emph{X buys a lottery ticket}, \textbf{xWant}, \textit{to win.}), we propose a \textbf{context, query, inference} (\textbf{CQI}) format with natural language queries serving as open relations. 
We also analyze unique properties of this distilled knowledge in  \S\ref{subsubsec:analysis}.

\subsubsection{Data Generation}
\label{subsubsec:data_gen}

We outline the generation process below, which consists of (1) generating contexts and (2) generating queries/inferences, resulting in our new knowledge base, \dataset.

\paragraph{Context Generation.}
First, we have experts generate 21 varied prompts to steer models to generate events or situations that require commonsense knowledge to fully understand (see \ref{sec:app-context} for all prompts used). 
As variations in prompt wording influence the model's output, we use many different prompts to enhance both diversity and coverage of the generated outputs. 
Half of the time, we generate contexts in a zero-shot manner, while for the other half, we do one-shot generation with one example drawn from \atomictenx \cite{west-etal-2022-symbolic}. In order to reduce generation cost, we generate 20 situations per prompt (batched generation).

We generate the contexts using GPT-3 \cite{brown2020language} variant \texttt{text-davinci-003} \cite{Ouyang2022TrainingLM}
for a total cost of
USD \$39.56. We set \texttt{top\_p=0.99} and \texttt{presence\_penalty=0.3},
lowering the logit values for tokens that have already occurred to promote diversity within each batch. Finally, to allow \model to see some diversity of names, we also randomly swap all entities (names or "PersonX/Y/Z") for a name drawn from the 2021 public US social security application name registry\footnote{\href{https://catalog.data.gov/dataset/baby-names-from-social-security-card-applications-national-data}{https://catalog.data.gov/dataset/baby-names-from-social-security-card-applications-national-data}} with probability 0.5.

\paragraph{Query/Inference Generation.} As no other resource currently has examples of high-quality commonsense inferences in our proposed open format, we develop a set of few-shot examples of 10 contexts (either handwritten or selected from \atomictenx or from ROCStories \cite{mostafazadeh-etal-2016-corpus}) with 10 handwritten commonsense query/inference pairs for each (see Appendix \ref{sec:app-inference} for more details). These query/inference pairs cover a broad swathe of knowledge, including consequences, explanations, reactions, attributes, counterfactuals, etc.

For each context in \data generated in the previous step, we randomly select and permute $n \sim \textrm{Uniform}(1,10)$ of the few-shot examples to provide in context after the instructions and then task the model with generating 10 query/inference pairs for the new context. 
The rationale for this random selection and permutation of the few-shot examples is to mitigate the potential overfitting to a specific ordering or selection or ordering of handwritten examples. 
To try to support the use case where a desired commonsense query is not known in advance, e.g. when a user simply want general knowledge for a given context, we also generated half of the commonsense hypotheses without generating a query first (prompt in \ref{sec:app-inference}). At training time (\S\ref{subsec:training}), we input a \texttt{NULL} value for the query field.
We generated all query/inference pairs using default decoding arguments with \texttt{gpt-3.5-turbo-0301} for a total cost of USD \$337.16.


\subsubsection{Analysis}
\label{subsubsec:analysis}

\begin{table}[t!]
\small
\centering 
    \begin{tabular}{l@{\hspace{0.6\tabcolsep}}|l@{\hspace{0.6\tabcolsep}}|r@{\hspace{0.6\tabcolsep}}r@{\hspace{0.6\tabcolsep}}|r@{\hspace{0.6\tabcolsep}}r@{\hspace{0.6\tabcolsep}}}
        \toprule 


            \multirow{2}{*}{\makecell[tl]{Type}} & \multirow{2}{*}{\makecell[tl]{Dataset \\ \textsc{Atomic}}} &  \multicolumn{2}{c|}{\textbf{Entries}} & \multicolumn{2}{c}{\textbf{3-grams}}  \\

             &  & \# & \% & \# & \% \\

            \midrule

            \multirow{3}{*}{\makecell[tl]{Context \\ \& \\ Event}} & \textsc{2020} & 43,958 & 3.5 & 40,194 & 55.8  \\ 

            & \textsc{10x} & 165,783 & 2.6 & 235,172 & 44.9 \\
            & \textsc{Nova} & 102,195 & 4.7 & 343,636 & 44.6 \\
            \midrule

            \multirow{3}{*}{\makecell[tl]{Query \\ \& \\ Relation}} & \textsc{2020} & 23 & - & - & - \\ 

            & \textsc{10x} & 7 & - & - & - \\
            & \textsc{Nova} & 822,615 & 79.2 & 1,609,780 & 28.7  \\
            \midrule

            \multirow{3}{*}{\makecell[tl]{Inference \\ \& \\ Tail}} & \textsc{2020} & 602,154 & 48.3 & 847,913 & 52.5 \\ 

            & \textsc{10x} & 874,417 & 13.5 & 695,877 & 21.0 \\
            & \textsc{Nova} & 2,030,488 & 93.2 & 5,835,099 & 30.0 \\
            \midrule

            \multirow{3}{*}{\makecell[tl]{Total}} & \textsc{2020} & 1,246,582 & - & 875,157 & 51.9  \\

            & \textsc{10x} & 6,456,300 & - & 812,166 & 21.1 \\
            & \textsc{Nova} & 2,178,086 & - & 7,224,608 & 28.0 \\





        \bottomrule
    \end{tabular}
    \caption{Statistics of \dataset compared to existing CSKG, \atomictwenty and \atomictenx. \textbf{\#} and \textbf{\%} indicate the count and percentage of unique entries or 3-grams, respectively. Compared to previous CSKGs, \dataset contains more diverse entries with higher lexical variations. Notably, as \dataset adopts open data format by breaking out from fixed relation types, it contains much more diverse and flexible sets of relations denoted with questions that tie premise and hypothesis together.}
\label{tab:data-comparison}
\end{table}


\paragraph{Comparison to Previous CSKGs.}

Table \ref{tab:data-comparison} shows the comparisons of \dataset to existing CSKGs, \atomictwenty \cite{Hwang2020COMETATOMIC2O} and \atomictenx \cite{west-etal-2022-symbolic} in dataset statistics and lexical diversity measures. \dataset contains more diverse unique premises (heads) and hypotheses (tails) as indicated by the higher number and/or percentage of these data entries.
\dataset also has higher lexical variations, as reflected by the significantly more diverse 3-grams.
In particular, as \dataset breaks out from fixed relation types with open questions to connect premise and hypothesis, it contains much more diverse and flexible sets of relations denoted by natural language questions.

It is also of note that, based on estimates from \cite{west-etal-2022-symbolic}, the total cost of \atomictwenty and \atomictenx were approximately USD \$40,000 and USD \$6,000 respectively, whereas the cost for \dataset is approximately \$400. Though the size of \dataset is somewhat smaller, the unit cost per example is also significantly lower. 

\paragraph{Analysis of Question Types.}

\begin{figure}
    \centering
    \includegraphics[width=0.95\linewidth]{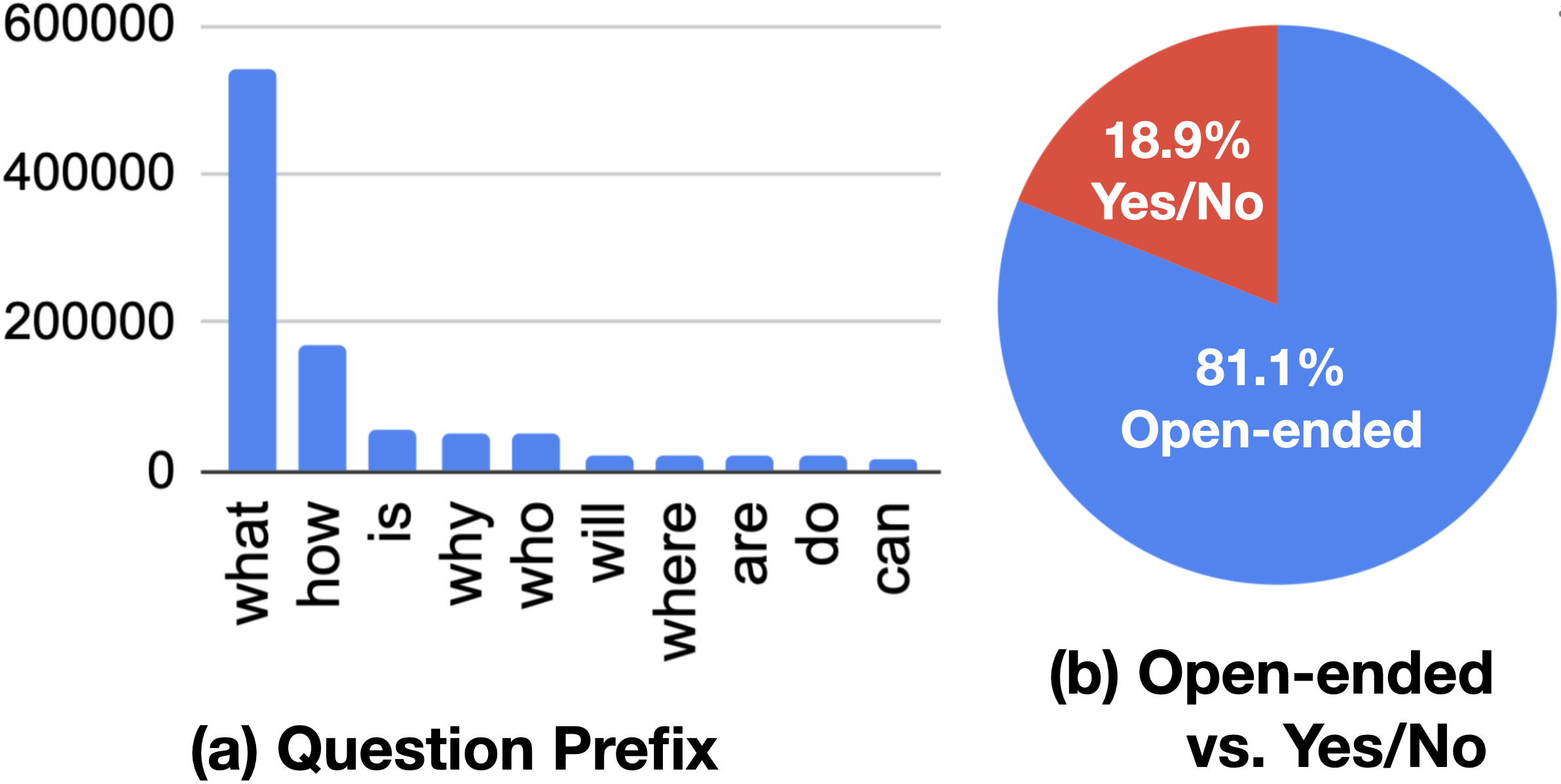}
    \caption{(a) The most frequent question prefixes. (b) The composition of open-ended vs. yes/no questions.}
    \label{fig:question-types}
\end{figure}

To delve into what relations are encoded in \dataset with open questions, we conduct an analysis of question types. Figure \ref{fig:question-types} shows the top 10 most common question prefixes, including open-ended question types, such as \textit{what} and \textit{how}, and binary yes/no question types, such as \textit{is} and \textit{will}. By grouping \textit{WH-questions} together (i.e., how, what, why, who, where, when, whose, which), we obtain 81.1\% of open-ended questions and 18.9\% of binary yes/no questions, indicating a diverse and flexible relation space the large portion of free-form questions represent, as shown in Figure \ref{tab:question-example}(b). Table \ref{tab:question-example} shows some of the most common questions in the dataset. The most common questions are not context-specific (asking about time, weather, or location), although we find that many of the queries do condition specifically on context.

\begin{table}[t!]
\small
\centering 
    \begin{tabular}{l@{\hspace{0.6\tabcolsep}}}
        \toprule 




            \textbf{Most Frequent Questions} \\
            \midrule
        
            What time is it? \\
            Who is PersonX? \\
            What is the weather like? \\
            What is the prerequisite for this situation? \\ 
            What is the consequence of the situation? \\ 
            What is the counterfactual of the situation? \\
            What will happen next? \\ 
            What is the occasion? \\ 
            What is the relationship between PersonX and PersonY? \\ 
            Where are they? \\
        \bottomrule
    \end{tabular}
    \caption{Frequent queries in \dataset. Note that we take the top 100 surface forms, and cluster them into semantically related/equivalent groups by hand. Queries above represent the top groups by aggregate count, with indicative labels. See Appendix~\ref{sec:app-queries} for more details.}
\label{tab:question-example}
\end{table}


\subsection{Plausibility Annotation}
\label{subsec:plaus_data}

Next, we collect annotations of CQI data plausibility. Broadly, this follows \citet{west-etal-2022-symbolic, Howard2023NeuroComparativesND} 
in combining generated data with an automatic critic to maximize the quality of a final trained model. In this case, however, we explore multiple ways to incorporate annotations of plausibility into the final model \model (\S\ref{subsubsec:versions}).

Our primary means of collecting annotations of plausibility is through Amazon Mechanical Turk. We use a similar annotation template to \cite{Hwang2020COMETATOMIC2O} (see Appendix~\ref{sec:app-amt}), asking annotators to decide if knowledge is \emph{always/often}, \emph{sometimes/likely}, \emph{farfetched/never} true, or \emph{invalid} (giving these annotations a score of 3, 2, 1, 0 respectively). 
We consider any knowledge scored 3 or 2 to be \emph{plausible}.

In practice, we collect 20k annotations, with 1 annotator per example. For underlying data, we use 16k examples from \dataset, as well as 2k examples each from \atomictenx and \atomictwenty to increase diversity of annotated knowledge style. While these knowledge graphs have fixed relation sets, we use sample natural language queries to replace the discrete relations (e.g. xNeed $\rightarrow$ \emph{What did PersonX need?}).

We conduct a human agreement study on a segment of 200 examples for which we elicit 3 annotations each, finding 
Fleiss $\kappa$ \cite{fleiss1971measuring} of 0.317 indicating fair agreement \cite{landis1977measurement}. 

\subsection{Training \model}
\label{subsec:training}

\subsubsection{Commonsense field masking}
\label{subsubsec:field_masking}

Previous knowledge models tend to use a standard \emph{head,relation} $\rightarrow$ \emph{tail} format in training, generating some inference given the situation/concept, and one of a set of possible commonsense relations to guide generation. 

The goal of \model is maximum flexibility in handling commonsense knowledge and tasks, meaning we would like to generate any of these fields from any others. For example, we may want to generate a likely \emph{query} that connects the \emph{context} and \emph{inference}; or, a context under which the query and inference are correct. To this end, we propose \emph{commonsense field masking}, wherein we randomly sample subspans of fields to be masked for prediction, e.g.

\begin{center}
\resizebox{.99\linewidth}{!}{
\begin{tabular}{l}
  \textbf{\texttt{Input: }} \\
  \textbf{\texttt{ \textcolor{headcolor}{Context: Consider the list of MASK$_{C}$ shows.}}} \\
  \textbf{\texttt{ \textcolor{relationcolor}{Query: What is the MASK$_{Q}$ show?}}} \\
  \textbf{\texttt{ \textcolor{tailcolor}{Inference: Hamilton}}}  \\
    \textbf{\texttt{Target: }} \\
  \textbf{\texttt{ \textcolor{headcolor}{MASK$_{C}$ = Broadway}}} \\
  \textbf{\texttt{ \textcolor{relationcolor}{MASK$_{Q}$ = most popular}}} 
\end{tabular}
} \end{center}

The process of masking follows two steps.  First, the set of which fields (CQI) will be masked is uniformly selected from all options in which at least one field is masked. Second, for each field, we randomly (with p=0.5) decide whether to mask the entire field, or a subspan. Finally, for those fields where a subspan is masked, we uniformly select the mask length, and which subspan of the given length to mask.

In effect, this gives the final model maximal flexibility at inference time. Users can mask any field, either the full span or infill a subspan as needed, allowing for use cases besides simply generating a full inference as in previous commonsense models. We explore how this can be especially useful in \S\ref{subsec:eval_gen}.

\subsubsection{\model Versions}
\label{subsubsec:versions}

We consider a variety of methods to use the generation and critique data described above for training.

\paragraph{Generation-only Model}
First, we consider the simplest option for producing a commonsense generation model: training directly on \dataset.
\modelgen is trained only on generation data from \S\ref{subsec:data} with the commonsense masking objective (\S\ref{subsubsec:field_masking}). Plausibility is not used in this model.

\paragraph{Critic-only Model}
Second, we train a stand-alone plausibility critic model, \modelcrit. This is trained to generate a plausibility score from a complete CQI (context, query, inference) knowledge triple, on the annotation set from \S\ref{subsec:plaus_data}. In effect, it returns a probability that a given CQI is plausible.

\paragraph{Filtered Generation Model}
Following \citet{west-etal-2022-symbolic}, we use a simple filtering-based technique for improving generation with plausibility scores. Using \modelcrit, we calculate the probability of being plausible for all examples in \dataset, and filter to only those points that meet a minimum probability. We focus on one threshold in particular, 0.99, indicating that \modelcrit gives at least 0.99 probability to the given CQI being plausible. We call the resulting model \modelfilter{0.99}, and the resulting filtered training set retains over 50\% of its original size, indicating \dataset is already high quality.


\paragraph{Quantized Reward Conditioning}
Inspired by quantized reward conditioning in \cite{Lu2022QuarkCT}, we also consider more closely unifying the critical and generation data. We consider a light-weight, one-step approach (as opposed to full reinforcement learning in \citealt{Lu2022QuarkCT}) in which we annotate \dataset with \modelcrit, then train a masked-prediction model that includes plausibility as a conditioning variable for predicting CQI. For annotation with \modelcrit, we greedily decode plausibility, and train a reward-conditional model \modelquarkg{}. When decoding with \modelquarkg{}, we condition on either of the ``plausible'' labels (2 or 3) from \S\ref{subsec:plaus_data}.

\subsubsection{Model Training}

We use the T5X codebase \citep{roberts2022t5x} to train \model, using the base T5 1.1 xxl ($\sim$11B parameters) checkpoint to initialize all of our experiments. We train all models on v3-128 TPU pods, using a batch size of 128 and a learning rate of 1e-5. For generation models, we train for a fixed 100k training steps, ensuring that loss does not converge or begin increasing. For models that include plausibility prediction as an objective, we stop training when evaluation loss for plausibility converges, which is often significantly before 100k training steps. 



\section{Experiments}
\label{sec:experiments}

\subsection{Evaluating Plausibility}
\label{subsec:eval_plaus}

We begin by evaluating the performance of our plausibility model \modelplaus. Particularly, we aim to understand the ability of this model to provide a useful, absolute plausibility score. We compare the accuracy of our plausibility scores on discriminative commonsense benchmarks to judge its effectiveness.

\subsubsection{Datasets}
We consider a range of standard discriminative commonsense benchmarks: HellaSwag (\textbf{HS}) \cite{Zellers2019HellaSwagCA} for generation recognition; \textbf{$\alpha$NLI} \cite{Bhagavatula2019AbductiveCR} for abductive reasoning; WinoGrande (\textbf{WG}) \cite{Sakaguchi2019WINOGRANDEAA} for pronoun resolution; Commonsense QA (\textbf{CSQA}) \cite{talmor-etal-2019-commonsenseqa} and \textbf{CODAH} \cite{chen2019codah} for general commonsense question answering; Social IQA (\textbf{SIQA}) \cite{Sap2019SocialIC} for social commonsense; RiddleSense (\textbf{RS}) \cite{lin-etal-2021-riddlesense} for riddle answering; and Physical IQA (\textbf{PIQA}) \cite{Bisk2019PIQARA} for physical commonsense. Together, these allow us to judge the ability of models to assess the correctness/plausibility of commonsense.

\subsubsection{Models and Baselines}

As baselines, we primarily consider popular language models in a roughly similar range of parameter sizes. We include basic language model LLaMA \cite{Touvron2023LLaMAOA} and PaLM \cite{Chowdhery2022PaLMSL} (citing performance directly for both); and language models with general task tuning such as QA for Macaw \cite{Tafjord2021Macaw} or instruction tuning for Flan-T5$_{xxl}$ \cite{Chung2022ScalingIL} and T0 \cite{Sanh2021MultitaskPT}. We create standard-format prompts that depend on the model. When possible, models are given answer choices as input. This is an advantage over plausibility models like \modelcrit which are designed to judge answers in isolation, but we include this to maximize baseline quality. To score options of baselines, we use negative-log-likelihood, as it was found by us to be best out of a range of options. We cite results for an alternative formatting for prompting FLAN from \cite{Liu2023VeraAG} which automatically reformats commonsense questions as statements, then judges plausibility as the likelihood of answering ``yes'' or ``no'' to whether the statement is plausible. We note that, while this method performs well, it will not generally apply to Context-Query-Inference (CQI) formatted data, as not all CQI examples can be naturally reformatted into short statements, but we include this baseline for completeness. We also cite results on GPT-3.5, ChatGPT, and GPT-4 from the same paper.

We compare baselines to \modelcrit described in \S\ref{subsec:training}. For this models, we score options based on the probability of predicting $2$ or $3$ for plausibility (sometimes/likely or always/often true), 
renormalized against the probability of predicting $1$ or $0$ (rarely or never true, or invalid). 

\begin{table*}[t!]
    \centering\footnotesize
     \scalebox{.98}{
    \begin{tabular}{l|cccccccccc}
\toprule
               system &  HS &  $\alpha$NLI &  CODAH &  WG &  CSQA &  SIQA &  CosmosQA &  RS &  PIQA \\
\midrule
\multicolumn{4}{l}{\textbf{Cited Results}} \\
\midrule
GPT-3.5 & 70.4 & 76.6  & 85  & 72.5 & 66.9 & 65.3 &-&-& 84.2 \\
ChatGPT & 43.0 & 60.3 & 56.8 & 61.3 & 39.6 & 52.2 &-&-& 67.6 \\
GPT-4 & 40.0 & 75.0 & 66.0 & 77.0 & 43.0 & 57.0 &-&-& 73.0 \\
\midrule
\midrule
   Flan-T5 (statements)$^1$ &       64.5 &  \textbf{80.8} &   \textbf{89.6} &        \textbf{84.7} &  69.2 &  \underline{73.2}  &  - & - &  \underline{83.9} \\
   llama-7B$^2$ & 76.1    &  -    &   -    &        70.1 &  -     &  48.9 &      -    & -   &  79.8 \\
   llama-13B$^2$ & 79.2    &  -    &   -    &        73.0 &  -     &  50.4 &      -    & -   &  80.1 \\
   PaLM 62B$^3$ &      \underline{79.7} &  - &  -  &        77.0 &  - &   - &      - &         - &  80.5 \\
   PaLM 540B$^3$ &       \textbf{83.4} &  - &  -  &        \underline{81.1} &  - &    - &      - &         - &  82.3 \\
   \midrule
   \multicolumn{4}{l}{\textbf{Comparable General Models}} \\
   \midrule
      Macaw &  50.8 &  71.6 &  82.9 &  60.7 &  \textbf{79.4} &  68.8 & 70.4 &  \underline{58.8} &  79.4 \\
      Flan-T5$_{xxl}$ & 73.5 &  70.7 &  58.7 &  72.9 & 72.8 &  55.2 &  72.9 &  \textbf{60.6} &  82.0 \\
      T0 & 63.7 &  70.3 &  73.4 &  58.9 &  68.1 &  66.8 &  \underline{75.4} &  53.8 &  \textbf{84.9} \\
   \midrule
   \midrule
   \modelcrit & 74.4 & \underline{80.4} & \underline{86.7} & 79.6 & \underline{76.7}  & \textbf{77.1} & \textbf{80.3} & 58.6 & 83.4 \\
\bottomrule
    \end{tabular}}
    \caption{ Comparison of model scores on commonsense benchmarks. \textbf{Best} results are bold and \underline{second} best are underlined. Note that no other method surpasses \modelcrit the combined number of 1st and 2nd place results (5). Comparison using \emph{absolute scores} from different models. $^1$ indicates values cited from \cite{Liu2023VeraAG} which uses a pipeline with Flan-T5$_{xxl}$, $^2$ indicates values cited from \cite{Chowdhery2022PaLMSL}, $^3$ indicates values cited from \cite{Touvron2023LLaMAOA}. Values for large, recent GPT models (GPT-3.5, ChatGPT, GPT4) are cited from \cite{Liu2023VeraAG}.
    }
    \label{tab:QA}
\end{table*}

\subsubsection{Results and Discussion}

Model scores on various tasks requiring commonsense knowledge can be seen in Table~\ref{tab:QA}. While various models are better at different tasks, \modelcrit is tied for most combined 1st + 2nd place results (5). Note that the other tied system, Flan-T5 (statements) requires automatically transforming each problem into a yes or no question; a transformation that is not generally applicable to the kinds of Context-Query-Inference style problems we would like to solve when deriving commonsense information from a model.

Looking at cases where \modelcrit fails to get the highest accuracy, it is perhaps unsurprising that PaLM 540B and 62B outperform all other models on HellaSwag, which requires predicting the most likely continuation of a scene description, a task especially well suited to a raw language model. Furthermore, with Physical IQA (PIQA), the focus is on \textit{physical} commonsense, a subcategory that our base generator seemed to produce less naturally on inspection.

We also note that many baselines (e.g. Macaw, T0) assume access to all answer choices. For our use case (judging knowledge within \dataset to improve the overall dataset) we are judging examples in isolation with no clear contrastive examples. The competitive performance of \modelcrit here, despite such disadvantages, further validates it for this use case.

\subsection{Evaluating Generation}
\label{subsec:eval_gen}

The central goal of \model is in generating commonsense knowledge, and carrying out commonsense reasoning. In this section, we test the ability of various versions of \model described in \S\ref{subsec:training} to do this. Note that we primarily use human evaluation for model generations, following a similar setup to \S\ref{subsec:plaus_data} with annotation templates available in Appendix \ref{sec:app-amt}.

\subsubsection{Datasets}

First, we test the ability of models to generate commonsense knowledge in the format of previous knowledge models. Particularly, we take a sample of \textbf{\atomictwenty} \cite{Hwang2020COMETATOMIC2O} commonsense prompts (\emph{head + relation}), testing the ability of models to generate a valid \emph{tail}. Results are included in Table~\ref{tab:downstream_generation}.

Next, we test on various downstream benchmarks requiring generative commonsense reasoning. First, we test abductive natural language generation (\textbf{$\alpha$NLG}) \cite{Bhagavatula2019AbductiveCR}, wherein models must abductively fill in the gap in a story between two observations. We also consider two question-answering datasets that require commonsense reasoning: \textbf{TellMeWhy} \cite{lal-etal-2021-tellmewhy} in which models explain events, and \textbf{Reflect} \cite{Zhou2022ReflectNR} in which models generate \textsc{Atomic}-style inferences for dialogues. We report results for all downstream reasoning benchmarks in Table~\ref{tab:QA}. We use a custom annotation template for $\alpha$NLG, and otherwise use the base CQI template from our annotation in \S\ref{subsec:plaus_data}.

\begin{table*}[t!]
    \centering\footnotesize
     \scalebox{.98}{
    \begin{tabular}{l|cccc|c|c|c}
\toprule
 & \multicolumn{4}{c|}{\textbf{$\alpha$NLG}} & \textbf{Reflect} & \textbf{TellMeWhy} & \textbf{\atomictwenty} \\
                                            system &  obs2 &  obs1 &  obs1+2 &  overall &  valid & valid & valid \\
\midrule
\multicolumn{2}{l}{\textbf{Baselines}} \\
\midrule
                                   LLaMA-7B &                  0.030 &                  0.025 &                     0.022 &                     0.013 &    0.388 & 0.463 &   0.470 \\
                                 LLaMA-13B &                  0.010 &                  0.008 &                     0.012 &                     0.008 &    0.456 & 0.442 &   0.515 \\
                                         T0 &                  0.260 &                  0.258 &                     0.235 &                     0.248 &    0.846 & 0.759 &   0.686 \\
                                  Alpaca-7b &                  0.162 &                  0.123 &                     0.120 &                     0.122 &    0.687 & 0.852 &   0.612 \\
                                 Alpaca-13B &                  0.355 &                  0.313 &                     0.290 &                     0.248 &    0.716 & 0.764 &   0.660 \\
                                   Flan-Ul2  &                  0.715 &                  0.627 &                     0.605 &                     0.622 &    0.618 & 0.562 &   0.692 \\
                                Flan-T5$_{xxl}$  &                  0.735 &                  0.653 &                     0.635 &                     0.657 &    0.796 & 0.807 &   0.757 \\
                                \midrule
                                \multicolumn{2}{l}{\textbf{\model}} \\
                                \midrule
\modelgen &                  0.877 &                  0.826 &                     0.819 &                     0.814 &    0.864 & \textbf{0.928} &   0.847 \\
\modelfilter{0.99} &                  \textbf{0.887} &                  \textbf{0.837} &                     \textbf{0.837} &                     \textbf{0.827} &    0.864 & 0.916 &   0.848 \\
          \modelquarkg{(2)} &                  0.837 &                  0.793 &                     0.787 &                     0.797 &    \textbf{0.874} & 0.916 &   \textbf{0.861} \\
          \modelquarkg{(3)} &                  0.840 &                  0.797 &                     0.780 &                     0.787 &    0.869 & 0.918 &   0.859 \\
\bottomrule
    \end{tabular}}
    \caption{Human evaluation of various commonsense generation tasks. Note that the basic version of \model outperforms baselines consistently, but is outperformed by versions that use plausibility to improve. We find human agreement with Fleiss $\kappa$ \cite{fleiss1971measuring} of 0.32, 0.44, 0.43, 0.39 (respective to order in the table) indicating fair to moderate agreement. Note, values in this table are normalized to a $[0, 1]$ range.
    }
    \label{tab:downstream_generation}
\end{table*}

\subsubsection{Baselines and Models}

For baselines, we include all of the models described in \S\ref{subsec:eval_plaus} as well as T5$_{xxl}$ ($\sim$11B parameters) finetuned for language modeling (\textbf{T5-LM}) \cite{Raffel2019ExploringTL}. We use straightforward prompts to describe each task and generate directly. 

Different datasets can demonstrate unique ways to use the commonsense masking of \model for generation. For example, for $\alpha$NLG, we mask between the beginning (\emph{o1}) and ending (\emph{o2}) events to form a natural sequence:

\begin{center}
\resizebox{.7\linewidth}{!}{
\begin{tabular}{l}
  \textbf{\texttt{Input: }} \\
  \textbf{\texttt{ \textcolor{headcolor}{Context: <o1> MASK$_{C}$ }}} \\
  \textbf{\texttt{ \textcolor{relationcolor}{Query: What happens next?}}} \\
  \textbf{\texttt{ \textcolor{tailcolor}{Inference: <o2>}}} 
\end{tabular}
} \end{center}

To predict a hypothesis \emph{h} that fits between \emph{o1} and \emph{o2}. We found this resulted in much higher quality generations than encoding  \emph{o1, o2} as context and predicting \emph{h} as inference.

For other datasets (Reflect, TellMeWhy, \atomictwenty), we can encode examples simply by giving context and query, then predicting the inference. 
For all models, we use greedy decoding.

\subsubsection{Results and Discussion}

All generation results use human evaluation, presented in Table~\ref{tab:downstream_generation}. Note that human evaluation templates are included in the Appendix. We evaluate 100 examples for each system and dataset. For Reflect, TellMeWhy, and \atomictwenty, we use the same template as \S\ref{subsec:plaus_data}. For $\alpha$NLG we use a template measuring coherence between the generated infill and either or both hypotheses, as well as overall quality. All scores in Table~\ref{tab:downstream_generation} are normalized to a range between 0 and 1.

Note that \model models win across the board. Particularly effective is the filtered model \modelfilter{0.99}, but so are the reward conditional models, and \modelquarkg{(2)} in particular, conditioned on ``2'' (likely/sometimes true) rather than ``3'' (always/often true). 
It is possible that answers that are always true are somewhat less creative or preferable to humans.

In general, the \model models that use plausibility information outperform the basic \modelgen, other than on the TellMeWhy dataset. This demonstrates a particular advantage of distilling discrete data -- it can be annotated, and those annotations can improve downstream performance.



Overall, superior performance of \model suggests that explicitly modeling knowledge can provide an advantage, at least considering tasks that explicitly require commonsense knowledge and reasoning.




\section{Related Work}
\label{sec:related_work}

\paragraph{Knowledge Generation}

Pretrained language models demonstrated the ability to carry implicit knowledge \cite{DBLP:journals/corr/abs-1909-01066, DBLP:journals/tacl/DhingraCEGEC22}.
These large language models are prompted for generating new knowledge to perform downstream tasks such as text classification \cite{DBLP:conf/emnlp/ShinRLWS20, DBLP:journals/corr/abs-1912-10165}, commonsense reasoning \cite{DBLP:conf/acl/0010LLWWBCH22, DBLP:journals/corr/abs-1806-02847, DBLP:conf/emnlp/DavisonFR19}. 
We take inspiration from commonsense LMs, designed for query commonsense knowledge, such as COMET \cite{bosselut2019comet} and COMET-2020 \cite{hwang2021comet}. 
Domain specific LMs are also used for knowledge graph completion in specialized domains like biomedicine \cite{DBLP:conf/akbc/NadkarniWBSHH21}. \citet{liu2022wanli} use dataset cartography to prime the model with challenging examples and enable it to generate more examples with such patterns.


\paragraph{Knowledge Distillation}
As the process of manually creating datasets can be costly and complex, prior studies have explored the realm of automated data generation. These prior works mainly focused on extractive approaches, e.g. syntactic parsing \cite{Zhang2020TransOMCSFL} or pattern matching \cite{causalbank} from unstructured text \cite{Lehmann2015DBpediaA, buck2014n}.

\citet{West2021SymbolicKD} proposed filtering out low quality data using a critic model for symbolic knowledge distillation from larger models. Following this, several works effectively improved upon this for iterative distillation \cite{sclar2022referee, Bhagavatula2022I2D2IK}, self-chat with feedback and conversations with ChatGPT \cite{Xu2023BaizeAO, koala_blogpost_2023, vicuna2023}. 
SODA \cite{kim2022soda} contextualized social commonsense knowledge from a knowledge graph to distill dialogues from InstructGPT. \citet{sclar2022referee} established filters based on length, fidelity, and information bottleneck for distilling reference-free summarization determining the effectiveness of designing filters for selecting data for the following iteration. 
Recently, \cite{Jung2023ImpossibleDF} proposed a framework to learn a high-quality model from a low-quality teacher model to distill a good dataset by summarizing and paraphrasing.


\section{Conclusions}
\label{sec:conclusions}

Overall, we introduce \model, an open commonsense foundation model. \model takes advantage of closed proprietary models, resulting in an open pipeline and resources that are publicly available. 
\model is trained on data generated from these closed proprietary models and augmented with human annotations, resulting in both a high-quality plausibility model and improved generative model. \model surpasses other general models of similar size at a range of commonsense knowledge-intensive tasks, demonstrating the existing need for explicit knowledge modeling, even as task-focused methods like instruction tuning grow in popularity.

\section*{Limitations}
First, we recognize that our line of research requires extensive resources and funding, limiting the broad adoption of our methodology as it is presented. Particularly, our work relies on both massive generation from proprietary language models (GPT-3 turbo) and extensive use of TPU resources. Our hope is that these barriers will only be lowered as proprietary LMs become cheaper and LMs become increasingly efficient to tune and do inference on \cite{dettmers2023qlora}, lowering the barrier for techniques such as ours.

Second of all, we recognize that, while we have attempted to test the query-ability of commonsense knowledge via automatic and human evaluations on a number of different tasks. However, current tasks are largely biased towards both certain topics and tends to implicitly define ground truth from certain, fixed perspectives rather than acknowledging the underlying diversity 
 of human perspectives \cite{santy2023nlpositionality}. This limits our ability to assess whether our models capture genuine human agreement—or only the agreement of a certain portion of the population—something which we hope future work will investigate.


\section*{Ethics Statement}

Akin to all other machine learning approaches, our model could inadvertently exhibit biases. We acknowledge that the open format relations gathered from proprietary models  may not be representative of all cultures, and thereby these perpetuate the biases that these proprietary large models possess. 
While generating commonsense knowledge, LLMs may result in unanticipated commonsense inferences, including those that are biased and escape our critic model. Consequently, incorporating these inferences during training can further amplify such biases. We are committed to understanding such biases and improving our critic model. 
However, our model's central tenet is knowledge, which contrasts with existing public models of similar size and architecture, 
thereby regulating the toxicity of the model.
We ensured that the crowd workers involved in our project were compensated at a rate that met or exceeded the minimum wage requirement, recognizing the value of their contributions to building our model.
Comparable to all open models, our model is susceptible to malicious use and it is our collective responsibility to thrust safe open usage.
We acutely understand the ethical implications associated with our proposed method and are dedicated to resolving them, aiming to ensure the responsible adaptation of our approach in the community.



 \bibliography{anthology,custom}

\begin{thebibliography}{58}
\expandafter\ifx\csname natexlab\endcsname\relax\def\natexlab#1{#1}\fi

\bibitem[{Bhagavatula et~al.(2023)Bhagavatula, Hwang, Downey, Bras, Lu, Sakaguchi, Swayamdipta, West, and Choi}]{Bhagavatula2022I2D2IK}
Chandra Bhagavatula, Jena~D. Hwang, Doug Downey, Ronan~Le Bras, Ximing Lu, Keisuke Sakaguchi, Swabha Swayamdipta, Peter West, and Yejin Choi. 2023.
\newblock {I}2{D}2: Inductive knowledge distillation with {N}euro{L}ogic and self-imitation.
\newblock In \emph{Proceedings of the 61st Annual Meeting of the Association for Computational Linguistics, {ACL}}.

\bibitem[{Bhagavatula et~al.(2019)Bhagavatula, {Le Bras}, Malaviya, Sakaguchi, Holtzman, Rashkin, Downey, Yih, and Choi}]{Bhagavatula2019AbductiveCR}
Chandra Bhagavatula, Ronan {Le Bras}, Chaitanya Malaviya, Keisuke Sakaguchi, Ari Holtzman, Hannah Rashkin, Doug Downey, Scott Yih, and Yejin Choi. 2019.
\newblock Abductive commonsense reasoning.
\newblock \emph{ICLR}.

\bibitem[{Bian et~al.(2023)Bian, Han, Sun, Lin, Lu, and He}]{Bian2023ChatGPTIA}
Ning Bian, Xianpei Han, Le~Sun, Hongyu Lin, Yaojie Lu, and Ben He. 2023.
\newblock Chatgpt is a knowledgeable but inexperienced solver: An investigation of commonsense problem in large language models.
\newblock \emph{ArXiv}, abs/2303.16421.

\bibitem[{Bisk et~al.(2019)Bisk, Zellers, Bras, Gao, and Choi}]{Bisk2019PIQARA}
Yonatan Bisk, Rowan Zellers, Ronan~Le Bras, Jianfeng Gao, and Yejin Choi. 2019.
\newblock {PIQA}: Reasoning about physical commonsense in natural language.
\newblock In \emph{AAAI Conference on Artificial Intelligence}.

\bibitem[{Bosselut et~al.(2019)Bosselut, Rashkin, Sap, Malaviya, Celikyilmaz, and Choi}]{bosselut2019comet}
Antoine Bosselut, Hannah Rashkin, Maarten Sap, Chaitanya Malaviya, Asli Celikyilmaz, and Yejin Choi. 2019.
\newblock Comet: Commonsense transformers for automatic knowledge graph construction.
\newblock In \emph{Proceedings of the 57th Annual Meeting of the Association for Computational Linguistics}, pages 4762--4779.

\bibitem[{Brown et~al.(2020)Brown, Mann, Ryder, Subbiah, Kaplan, Dhariwal, Neelakantan, Shyam, Sastry, Askell, Agarwal, Herbert-Voss, Krueger, Henighan, Child, Ramesh, Ziegler, Wu, Winter, Hesse, Chen, Sigler, Litwin, Gray, Chess, Clark, Berner, McCandlish, Radford, Sutskever, and Amodei}]{brown2020language}
Tom~B. Brown, Benjamin Mann, Nick Ryder, Melanie Subbiah, Jared Kaplan, Prafulla Dhariwal, Arvind Neelakantan, Pranav Shyam, Girish Sastry, Amanda Askell, Sandhini Agarwal, Ariel Herbert-Voss, Gretchen Krueger, Tom Henighan, Rewon Child, Aditya Ramesh, Daniel~M. Ziegler, Jeffrey Wu, Clemens Winter, Christopher Hesse, Mark Chen, Eric Sigler, Mateusz Litwin, Scott Gray, Benjamin Chess, Jack Clark, Christopher Berner, Sam McCandlish, Alec Radford, Ilya Sutskever, and Dario Amodei. 2020.
\newblock \href {http://arxiv.org/abs/2005.14165} {Language models are few-shot learners}.

\bibitem[{Bubeck et~al.(2023)Bubeck, Chandrasekaran, Eldan, Gehrke, Horvitz, Kamar, Lee, Lee, Li, Lundberg, Nori, Palangi, Ribeiro, and Zhang}]{bubeck2023sparks}
Sébastien Bubeck, Varun Chandrasekaran, Ronen Eldan, Johannes Gehrke, Eric Horvitz, Ece Kamar, Peter Lee, Yin~Tat Lee, Yuanzhi Li, Scott Lundberg, Harsha Nori, Hamid Palangi, Marco~Tulio Ribeiro, and Yi~Zhang. 2023.
\newblock \href {http://arxiv.org/abs/2303.12712} {Sparks of artificial general intelligence: Early experiments with gpt-4}.

\bibitem[{Buck et~al.(2014)Buck, Heafield, and Van~Ooyen}]{buck2014n}
Christian Buck, Kenneth Heafield, and Bas Van~Ooyen. 2014.
\newblock N-gram counts and language models from the common crawl.
\newblock In \emph{LREC}, volume~2, page~4. Citeseer.

\bibitem[{Chen et~al.(2019)Chen, D'Arcy, Liu, Fernandez, and Downey}]{chen2019codah}
Michael Chen, Mike D'Arcy, Alisa Liu, Jared Fernandez, and Doug Downey. 2019.
\newblock Codah: An adversarially-authored question answering dataset for common sense.
\newblock In \emph{Proceedings of the 3rd Workshop on Evaluating Vector Space Representations for NLP}, pages 63--69.

\bibitem[{Chiang et~al.(2023)Chiang, Li, Lin, Sheng, Wu, Zhang, Zheng, Zhuang, Zhuang, Gonzalez, Stoica, and Xing}]{vicuna2023}
Wei-Lin Chiang, Zhuohan Li, Zi~Lin, Ying Sheng, Zhanghao Wu, Hao Zhang, Lianmin Zheng, Siyuan Zhuang, Yonghao Zhuang, Joseph~E. Gonzalez, Ion Stoica, and Eric~P. Xing. 2023.
\newblock \href {https://lmsys.org/blog/2023-03-30-vicuna/} {Vicuna: An open-source chatbot impressing gpt-4 with 90\%* chatgpt quality}.

\bibitem[{Chowdhery et~al.(2022)Chowdhery, Narang, Devlin, Bosma, Mishra, Roberts, Barham, Chung, Sutton, Gehrmann, Schuh, Shi, Tsvyashchenko, Maynez, Rao, Barnes, Tay, Shazeer, Prabhakaran, Reif, Du, Hutchinson, Pope, Bradbury, Austin, Isard, Gur-Ari, Yin, Duke, Levskaya, Ghemawat, Dev, Michalewski, Garc{\'i}a, Misra, Robinson, Fedus, Zhou, Ippolito, Luan, Lim, Zoph, Spiridonov, Sepassi, Dohan, Agrawal, Omernick, Dai, Pillai, Pellat, Lewkowycz, Moreira, Child, Polozov, Lee, Zhou, Wang, Saeta, D{\'i}az, Firat, Catasta, Wei, Meier-Hellstern, Eck, Dean, Petrov, and Fiedel}]{Chowdhery2022PaLMSL}
Aakanksha Chowdhery, Sharan Narang, Jacob Devlin, Maarten Bosma, Gaurav Mishra, Adam Roberts, Paul Barham, Hyung~Won Chung, Charles Sutton, Sebastian Gehrmann, Parker Schuh, Kensen Shi, Sasha Tsvyashchenko, Joshua Maynez, Abhishek Rao, Parker Barnes, Yi~Tay, Noam~M. Shazeer, Vinodkumar Prabhakaran, Emily Reif, Nan Du, Benton~C. Hutchinson, Reiner Pope, James Bradbury, Jacob Austin, Michael Isard, Guy Gur-Ari, Pengcheng Yin, Toju Duke, Anselm Levskaya, Sanjay Ghemawat, Sunipa Dev, Henryk Michalewski, Xavier Garc{\'i}a, Vedant Misra, Kevin Robinson, Liam Fedus, Denny Zhou, Daphne Ippolito, David Luan, Hyeontaek Lim, Barret Zoph, Alexander Spiridonov, Ryan Sepassi, David Dohan, Shivani Agrawal, Mark Omernick, Andrew~M. Dai, Thanumalayan~Sankaranarayana Pillai, Marie Pellat, Aitor Lewkowycz, Erica Moreira, Rewon Child, Oleksandr Polozov, Katherine Lee, Zongwei Zhou, Xuezhi Wang, Brennan Saeta, Mark D{\'i}az, Orhan Firat, Michele Catasta, Jason Wei, Kathleen~S. Meier-Hellstern, Douglas Eck, Jeff Dean, Slav Petrov,
  and Noah Fiedel. 2022.
\newblock Palm: Scaling language modeling with pathways.
\newblock \emph{ArXiv}, abs/2204.02311.

\bibitem[{Chung et~al.(2022{\natexlab{a}})Chung, Hou, Longpre, Zoph, Tay, Fedus, Li, Wang, Dehghani, Brahma, Webson, Gu, Dai, Suzgun, Chen, Chowdhery, Narang, Mishra, Yu, Zhao, Huang, Dai, Yu, Petrov, Chi, Dean, Devlin, Roberts, Zhou, Le, and Wei}]{Chung2022flan}
Hyung~Won Chung, Le~Hou, S.~Longpre, Barret Zoph, Yi~Tay, William Fedus, Eric Li, Xuezhi Wang, Mostafa Dehghani, Siddhartha Brahma, Albert Webson, Shixiang~Shane Gu, Zhuyun Dai, Mirac Suzgun, Xinyun Chen, Aakanksha Chowdhery, Sharan Narang, Gaurav Mishra, Adams~Wei Yu, Vincent Zhao, Yanping Huang, Andrew~M. Dai, Hongkun Yu, Slav Petrov, Ed~Chi, Jeff Dean, Jacob Devlin, Adam Roberts, Denny Zhou, Quoc Le, and Jason Wei. 2022{\natexlab{a}}.
\newblock Scaling instruction-finetuned language models.
\newblock \emph{ArXiv}, abs/2210.11416.

\bibitem[{Chung et~al.(2022{\natexlab{b}})Chung, Hou, Longpre, Zoph, Tay, Fedus, Li, Wang, Dehghani, Brahma, Webson, Gu, Dai, Suzgun, Chen, Chowdhery, Valter, Narang, Mishra, Yu, Zhao, Huang, Dai, Yu, Petrov, hsin Chi, Dean, Devlin, Roberts, Zhou, Le, and Wei}]{Chung2022ScalingIL}
Hyung~Won Chung, Le~Hou, S.~Longpre, Barret Zoph, Yi~Tay, William Fedus, Eric Li, Xuezhi Wang, Mostafa Dehghani, Siddhartha Brahma, Albert Webson, Shixiang~Shane Gu, Zhuyun Dai, Mirac Suzgun, Xinyun Chen, Aakanksha Chowdhery, Dasha Valter, Sharan Narang, Gaurav Mishra, Adams~Wei Yu, Vincent Zhao, Yanping Huang, Andrew~M. Dai, Hongkun Yu, Slav Petrov, Ed~Huai hsin Chi, Jeff Dean, Jacob Devlin, Adam Roberts, Denny Zhou, Quoc~V. Le, and Jason Wei. 2022{\natexlab{b}}.
\newblock Scaling instruction-finetuned language models.
\newblock \emph{ArXiv}, abs/2210.11416.

\bibitem[{Davison et~al.(2019)Davison, Feldman, and Rush}]{DBLP:conf/emnlp/DavisonFR19}
Joe Davison, Joshua Feldman, and Alexander~M. Rush. 2019.
\newblock \href {https://doi.org/10.18653/v1/D19-1109} {Commonsense knowledge mining from pretrained models}.
\newblock In \emph{Proceedings of the 2019 Conference on Empirical Methods in Natural Language Processing and the 9th International Joint Conference on Natural Language Processing, {EMNLP-IJCNLP} 2019, Hong Kong, China, November 3-7, 2019}, pages 1173--1178. Association for Computational Linguistics.

\bibitem[{Dettmers et~al.(2023)Dettmers, Pagnoni, Holtzman, and Zettlemoyer}]{dettmers2023qlora}
Tim Dettmers, Artidoro Pagnoni, Ari Holtzman, and Luke Zettlemoyer. 2023.
\newblock Qlora: Efficient finetuning of quantized llms.
\newblock \emph{arXiv preprint arXiv:2305.14314}.

\bibitem[{Dhingra et~al.(2022)Dhingra, Cole, Eisenschlos, Gillick, Eisenstein, and Cohen}]{DBLP:journals/tacl/DhingraCEGEC22}
Bhuwan Dhingra, Jeremy~R. Cole, Julian~Martin Eisenschlos, Daniel Gillick, Jacob Eisenstein, and William~W. Cohen. 2022.
\newblock \href {https://doi.org/10.1162/tacl\_a\_00459} {Time-aware language models as temporal knowledge bases}.
\newblock \emph{Trans. Assoc. Comput. Linguistics}, 10:257--273.

\bibitem[{Fleiss(1971)}]{fleiss1971measuring}
Joseph~L Fleiss. 1971.
\newblock Measuring nominal scale agreement among many raters.
\newblock \emph{Psychological bulletin}, 76(5):378.

\bibitem[{Geng et~al.(2023)Geng, Gudibande, Liu, Wallace, Abbeel, Levine, and Song}]{koala_blogpost_2023}
Xinyang Geng, Arnav Gudibande, Hao Liu, Eric Wallace, Pieter Abbeel, Sergey Levine, and Dawn Song. 2023.
\newblock \href {https://bair.berkeley.edu/blog/2023/04/03/koala/} {Koala: A dialogue model for academic research}.
\newblock Blog post.

\bibitem[{Howard et~al.(2023)Howard, Wang, Lal, Singer, Choi, and Swayamdipta}]{Howard2023NeuroComparativesND}
Phillip Howard, Junlin Wang, Vasudev Lal, Gadi Singer, Yejin Choi, and Swabha Swayamdipta. 2023.
\newblock Neurocomparatives: Neuro-symbolic distillation of comparative knowledge.
\newblock \emph{ArXiv}, abs/2305.04978.

\bibitem[{Hwang et~al.(2020)Hwang, Bhagavatula, {Le Bras}, Da, Sakaguchi, Bosselut, and Choi}]{Hwang2020COMETATOMIC2O}
Jena~D. Hwang, Chandra Bhagavatula, Ronan {Le Bras}, Jeff Da, Keisuke Sakaguchi, Antoine Bosselut, and Yejin Choi. 2020.
\newblock Comet-atomic 2020: On symbolic and neural commonsense knowledge graphs.
\newblock In \emph{AAAI Conference on Artificial Intelligence}.

\bibitem[{Hwang et~al.(2021)Hwang, Bhagavatula, Le~Bras, Da, Sakaguchi, Bosselut, and Choi}]{hwang2021comet}
Jena~D Hwang, Chandra Bhagavatula, Ronan Le~Bras, Jeff Da, Keisuke Sakaguchi, Antoine Bosselut, and Yejin Choi. 2021.
\newblock (comet-) atomic 2020: On symbolic and neural commonsense knowledge graphs.
\newblock In \emph{Proceedings of the AAAI Conference on Artificial Intelligence}, volume~35, pages 6384--6392.

\bibitem[{Jung et~al.(2023)Jung, West, Jiang, Brahman, Lu, Fisher, Sorensen, and Choi}]{Jung2023ImpossibleDF}
Jaehun Jung, Peter West, Liwei Jiang, Faeze Brahman, Ximing Lu, Jillian Fisher, Taylor Sorensen, and Yejin Choi. 2023.
\newblock Impossible distillation: from low-quality model to high-quality dataset \& model for summarization and paraphrasing.
\newblock \emph{ArXiv}, abs/2305.16635.

\bibitem[{Kim et~al.(2023)Kim, Hessel, Jiang, Lu, Yu, Zhou, Le~Bras, Alikhani, Kim, Sap et~al.}]{kim2022soda}
Hyunwoo Kim, Jack Hessel, Liwei Jiang, Ximing Lu, Youngjae Yu, Pei Zhou, Ronan Le~Bras, Malihe Alikhani, Gunhee Kim, Maarten Sap, et~al. 2023.
\newblock {S}{O}{D}{A}: Million-scale dialogue distillation with social commonsense contextualization.
\newblock In \emph{Proceedings of the 2023 Conference on Empirical Methods in Natural Language Processing, {EMNLP}}.

\bibitem[{Lal et~al.(2021)Lal, Chambers, Mooney, and Balasubramanian}]{lal-etal-2021-tellmewhy}
Yash~Kumar Lal, Nathanael Chambers, Raymond Mooney, and Niranjan Balasubramanian. 2021.
\newblock \href {https://doi.org/10.18653/v1/2021.findings-acl.53} {{T}ell{M}e{W}hy: A dataset for answering why-questions in narratives}.
\newblock In \emph{Findings of the Association for Computational Linguistics: ACL-IJCNLP 2021}, pages 596--610, Online. Association for Computational Linguistics.

\bibitem[{Landis and Koch(1977)}]{landis1977measurement}
J~Richard Landis and Gary~G Koch. 1977.
\newblock The measurement of observer agreement for categorical data.
\newblock \emph{biometrics}, pages 159--174.

\bibitem[{Lehmann et~al.(2015)Lehmann, Isele, Jakob, Jentzsch, Kontokostas, Mendes, Hellmann, Morsey, van Kleef, Auer, and Bizer}]{Lehmann2015DBpediaA}
Jens Lehmann, Robert Isele, Max Jakob, Anja Jentzsch, D.~Kontokostas, Pablo~N. Mendes, Sebastian Hellmann, M.~Morsey, Patrick van Kleef, S.~Auer, and C.~Bizer. 2015.
\newblock Dbpedia - a large-scale, multilingual knowledge base extracted from wikipedia.
\newblock \emph{Semantic Web}, 6:167--195.

\bibitem[{Li et~al.(2020)Li, Ding, Liu, Hu, and Van~Durme}]{causalbank}
Zhongyang Li, Xiao Ding, Ting Liu, J.~Edward Hu, and Benjamin Van~Durme. 2020.
\newblock Guided generation of cause and effect.
\newblock In \emph{Proceedings of the Twenty-Ninth International Joint Conference on Artificial Intelligence, {IJCAI-20}}.

\bibitem[{Lin et~al.(2021)Lin, Wu, Yang, Lee, and Ren}]{lin-etal-2021-riddlesense}
Bill~Yuchen Lin, Ziyi Wu, Yichi Yang, Dong-Ho Lee, and Xiang Ren. 2021.
\newblock \href {https://doi.org/10.18653/v1/2021.findings-acl.131} {{R}iddle{S}ense: Reasoning about riddle questions featuring linguistic creativity and commonsense knowledge}.
\newblock In \emph{Findings of the Association for Computational Linguistics: ACL-IJCNLP 2021}, pages 1504--1515, Online. Association for Computational Linguistics.

\bibitem[{Liu et~al.(2022{\natexlab{a}})Liu, Swayamdipta, Smith, and Choi}]{liu2022wanli}
Alisa Liu, Swabha Swayamdipta, Noah~A. Smith, and Yejin Choi. 2022{\natexlab{a}}.
\newblock \href {https://api.semanticscholar.org/CorpusID:246016339} {{W}{A}{N}{L}{I}: Worker and ai collaboration for natural language inference dataset creation}.
\newblock In \emph{Conference on Empirical Methods in Natural Language Processing}.

\bibitem[{Liu et~al.(2022{\natexlab{b}})Liu, Liu, Lu, Welleck, West, {Le Bras}, Choi, and Hajishirzi}]{DBLP:conf/acl/0010LLWWBCH22}
Jiacheng Liu, Alisa Liu, Ximing Lu, Sean Welleck, Peter West, Ronan {Le Bras}, Yejin Choi, and Hannaneh Hajishirzi. 2022{\natexlab{b}}.
\newblock \href {https://doi.org/10.18653/v1/2022.acl-long.225} {Generated knowledge prompting for commonsense reasoning}.
\newblock In \emph{Proceedings of the 60th Annual Meeting of the Association for Computational Linguistics (Volume 1: Long Papers), {ACL} 2022, Dublin, Ireland, May 22-27, 2022}, pages 3154--3169. Association for Computational Linguistics.

\bibitem[{Liu et~al.(2023)Liu, Wang, Wang, Smith, Choi, and Hajishirzi}]{Liu2023VeraAG}
Jiacheng Liu, Wenya Wang, Dianzhuo Wang, Noah~A. Smith, Yejin Choi, and Hanna Hajishirzi. 2023.
\newblock Vera: A general-purpose plausibility estimation model for commonsense statements.
\newblock \emph{ArXiv}, abs/2305.03695.

\bibitem[{Lu et~al.(2022)Lu, Welleck, Hessel, Jiang, Qin, West, Ammanabrolu, and Choi}]{Lu2022QuarkCT}
Ximing Lu, Sean Welleck, Jack Hessel, Liwei Jiang, Lianhui Qin, Peter West, Prithviraj Ammanabrolu, and Yejin Choi. 2022.
\newblock \href {https://openreview.net/forum?id=5HaIds3ux5O} {{QUARK}: Controllable text generation with reinforced unlearning}.
\newblock In \emph{Advances in Neural Information Processing Systems}.

\bibitem[{Mostafazadeh et~al.(2016)Mostafazadeh, Chambers, He, Parikh, Batra, Vanderwende, Kohli, and Allen}]{mostafazadeh-etal-2016-corpus}
Nasrin Mostafazadeh, Nathanael Chambers, Xiaodong He, Devi Parikh, Dhruv Batra, Lucy Vanderwende, Pushmeet Kohli, and James Allen. 2016.
\newblock \href {https://doi.org/10.18653/v1/N16-1098} {A corpus and cloze evaluation for deeper understanding of commonsense stories}.
\newblock In \emph{Proceedings of the 2016 Conference of the North {A}merican Chapter of the Association for Computational Linguistics: Human Language Technologies}, pages 839--849, San Diego, California. Association for Computational Linguistics.

\bibitem[{Nadkarni et~al.(2021)Nadkarni, Wadden, Beltagy, Smith, Hajishirzi, and Hope}]{DBLP:conf/akbc/NadkarniWBSHH21}
Rahul Nadkarni, David Wadden, Iz~Beltagy, Noah~A. Smith, Hannaneh Hajishirzi, and Tom Hope. 2021.
\newblock \href {https://doi.org/10.24432/C5QC7V} {Scientific language models for biomedical knowledge base completion: An empirical study}.
\newblock In \emph{3rd Conference on Automated Knowledge Base Construction, {AKBC} 2021, Virtual, October 4-8, 2021}.

\bibitem[{OpenAI(2023)}]{OpenAI2023GPT4TR}
OpenAI. 2023.
\newblock Gpt-4 technical report.
\newblock \emph{ArXiv}, abs/2303.08774.

\bibitem[{Ouyang et~al.(2022)Ouyang, Wu, Jiang, Almeida, Wainwright, Mishkin, Zhang, Agarwal, Slama, Ray, Schulman, Hilton, Kelton, Miller, Simens, Askell, Welinder, Christiano, Leike, and Lowe}]{Ouyang2022TrainingLM}
Long Ouyang, Jeff Wu, Xu~Jiang, Diogo Almeida, Carroll~L. Wainwright, Pamela Mishkin, Chong Zhang, Sandhini Agarwal, Katarina Slama, Alex Ray, John Schulman, Jacob Hilton, Fraser Kelton, Luke~E. Miller, Maddie Simens, Amanda Askell, Peter Welinder, Paul~Francis Christiano, Jan Leike, and Ryan~J. Lowe. 2022.
\newblock Training language models to follow instructions with human feedback.
\newblock \emph{ArXiv}, abs/2203.02155.

\bibitem[{Papineni et~al.(2002)Papineni, Roukos, Ward, and Zhu}]{papineni2002bleu}
Kishore Papineni, Salim Roukos, Todd Ward, and Wei-Jing Zhu. 2002.
\newblock Bleu: a method for automatic evaluation of machine translation.
\newblock In \emph{Proceedings of the 40th annual meeting on association for computational linguistics}, pages 311--318. Association for Computational Linguistics.

\bibitem[{Petroni et~al.(2019)Petroni, Rockt{\"{a}}schel, Lewis, Bakhtin, Wu, Miller, and Riedel}]{DBLP:journals/corr/abs-1909-01066}
Fabio Petroni, Tim Rockt{\"{a}}schel, Patrick S.~H. Lewis, Anton Bakhtin, Yuxiang Wu, Alexander~H. Miller, and Sebastian Riedel. 2019.
\newblock \href {http://arxiv.org/abs/1909.01066} {Language models as knowledge bases?}
\newblock \emph{CoRR}, abs/1909.01066.

\bibitem[{Puri and Catanzaro(2019)}]{DBLP:journals/corr/abs-1912-10165}
Raul Puri and Bryan Catanzaro. 2019.
\newblock \href {http://arxiv.org/abs/1912.10165} {Zero-shot text classification with generative language models}.
\newblock \emph{CoRR}, abs/1912.10165.

\bibitem[{Raffel et~al.(2019)Raffel, Shazeer, Roberts, Lee, Narang, Matena, Zhou, Li, and Liu}]{Raffel2019ExploringTL}
Colin Raffel, Noam~M. Shazeer, Adam Roberts, Katherine Lee, Sharan Narang, Michael Matena, Yanqi Zhou, Wei Li, and Peter~J. Liu. 2019.
\newblock Exploring the limits of transfer learning with a unified text-to-text transformer.
\newblock \emph{ArXiv}, abs/1910.10683.

\bibitem[{Roberts et~al.(2022)Roberts, Chung, Levskaya, Mishra, Bradbury, Andor, Narang, Lester, Gaffney, Mohiuddin, Hawthorne, Lewkowycz, Salcianu, van Zee, Austin, Goodman, Soares, Hu, Tsvyashchenko, Chowdhery, Bastings, Bulian, Garcia, Ni, Chen, Kenealy, Clark, Lee, Garrette, Lee-Thorp, Raffel, Shazeer, Ritter, Bosma, Passos, Maitin-Shepard, Fiedel, Omernick, Saeta, Sepassi, Spiridonov, Newlan, and Gesmundo}]{roberts2022t5x}
Adam Roberts, Hyung~Won Chung, Anselm Levskaya, Gaurav Mishra, James Bradbury, Daniel Andor, Sharan Narang, Brian Lester, Colin Gaffney, Afroz Mohiuddin, Curtis Hawthorne, Aitor Lewkowycz, Alex Salcianu, Marc van Zee, Jacob Austin, Sebastian Goodman, Livio~Baldini Soares, Haitang Hu, Sasha Tsvyashchenko, Aakanksha Chowdhery, Jasmijn Bastings, Jannis Bulian, Xavier Garcia, Jianmo Ni, Andrew Chen, Kathleen Kenealy, Jonathan~H. Clark, Stephan Lee, Dan Garrette, James Lee-Thorp, Colin Raffel, Noam Shazeer, Marvin Ritter, Maarten Bosma, Alexandre Passos, Jeremy Maitin-Shepard, Noah Fiedel, Mark Omernick, Brennan Saeta, Ryan Sepassi, Alexander Spiridonov, Joshua Newlan, and Andrea Gesmundo. 2022.
\newblock \href {https://arxiv.org/abs/2203.17189} {Scaling up models and data with $\texttt{t5x}$ and $\texttt{seqio}$}.
\newblock \emph{arXiv preprint arXiv:2203.17189}.

\bibitem[{Sakaguchi et~al.(2019)Sakaguchi, {Le Bras}, Bhagavatula, and Choi}]{Sakaguchi2019WINOGRANDEAA}
Keisuke Sakaguchi, Ronan {Le Bras}, Chandra Bhagavatula, and Yejin Choi. 2019.
\newblock Winogrande: An adversarial winograd schema challenge at scale.
\newblock In \emph{AAAI Conference on Artificial Intelligence}.

\bibitem[{Sanh et~al.(2021)Sanh, Webson, Raffel, Bach, Sutawika, Alyafeai, Chaffin, Stiegler, Scao, Raja, Dey, Bari, Xu, Thakker, Sharma, Szczechla, Kim, Chhablani, Nayak, Datta, Chang, Jiang, Wang, Manica, Shen, Yong, Pandey, Bawden, Wang, Neeraj, Rozen, Sharma, Santilli, F{\'e}vry, Fries, Teehan, Biderman, Gao, Bers, Wolf, and Rush}]{Sanh2021MultitaskPT}
Victor Sanh, Albert Webson, Colin Raffel, Stephen~H. Bach, Lintang Sutawika, Zaid Alyafeai, Antoine Chaffin, Arnaud Stiegler, Teven~Le Scao, Arun Raja, Manan Dey, M~Saiful Bari, Canwen Xu, Urmish Thakker, Shanya Sharma, Eliza Szczechla, Taewoon Kim, Gunjan Chhablani, Nihal~V. Nayak, Debajyoti Datta, Jonathan Chang, Mike Tian-Jian Jiang, Han Wang, Matteo Manica, Sheng Shen, Zheng~Xin Yong, Harshit Pandey, Rachel Bawden, Thomas Wang, Trishala Neeraj, Jos Rozen, Abheesht Sharma, Andrea Santilli, Thibault F{\'e}vry, Jason~Alan Fries, Ryan Teehan, Stella~Rose Biderman, Leo Gao, Tali Bers, Thomas Wolf, and Alexander~M. Rush. 2021.
\newblock Multitask prompted training enables zero-shot task generalization.
\newblock \emph{ArXiv}, abs/2110.08207.

\bibitem[{Santy et~al.(2023)Santy, Liang, Bras, Reinecke, and Sap}]{santy2023nlpositionality}
Sebastin Santy, Jenny~T Liang, Ronan~Le Bras, Katharina Reinecke, and Maarten Sap. 2023.
\newblock {NLP}ositionality: Characterizing design biases of datasets and models.
\newblock In \emph{Proceedings of the 61st Annual Meeting of the Association for Computational Linguistics, {ACL}}.

\bibitem[{Sap et~al.(2019)Sap, Rashkin, Chen, Bras, and Choi}]{Sap2019SocialIC}
Maarten Sap, Hannah Rashkin, Derek Chen, Ronan~Le Bras, and Yejin Choi. 2019.
\newblock Social {IQA}: Commonsense reasoning about social interactions.
\newblock In \emph{Conference on Empirical Methods in Natural Language Processing}.

\bibitem[{Sclar et~al.(2022)Sclar, West, Kumar, Tsvetkov, and Choi}]{sclar2022referee}
Melanie Sclar, Peter West, Sachin Kumar, Yulia Tsvetkov, and Yejin Choi. 2022.
\newblock Referee: Reference-free sentence summarization with sharper controllability through symbolic knowledge distillation.
\newblock \emph{arXiv preprint arXiv:2210.13800}.

\bibitem[{Shin et~al.(2020)Shin, Razeghi, IV, Wallace, and Singh}]{DBLP:conf/emnlp/ShinRLWS20}
Taylor Shin, Yasaman Razeghi, Robert L.~Logan IV, Eric Wallace, and Sameer Singh. 2020.
\newblock \href {https://doi.org/10.18653/v1/2020.emnlp-main.346} {Autoprompt: Eliciting knowledge from language models with automatically generated prompts}.
\newblock In \emph{Proceedings of the 2020 Conference on Empirical Methods in Natural Language Processing, {EMNLP} 2020, Online, November 16-20, 2020}, pages 4222--4235. Association for Computational Linguistics.

\bibitem[{Tafjord and Clark(2021)}]{Tafjord2021Macaw}
Oyvind Tafjord and Peter Clark. 2021.
\newblock General-purpose question-answering with macaw.
\newblock \emph{ArXiv}, abs/2109.02593.

\bibitem[{Talmor et~al.(2019)Talmor, Herzig, Lourie, and Berant}]{talmor-etal-2019-commonsenseqa}
Alon Talmor, Jonathan Herzig, Nicholas Lourie, and Jonathan Berant. 2019.
\newblock \href {https://doi.org/10.18653/v1/N19-1421} {{C}ommonsense{QA}: A question answering challenge targeting commonsense knowledge}.
\newblock In \emph{Proceedings of the 2019 Conference of the North {A}merican Chapter of the Association for Computational Linguistics: Human Language Technologies, Volume 1 (Long and Short Papers)}, pages 4149--4158, Minneapolis, Minnesota. Association for Computational Linguistics.

\bibitem[{Touvron et~al.(2023)Touvron, Lavril, Izacard, Martinet, Lachaux, Lacroix, Rozi{\`e}re, Goyal, Hambro, Azhar, Rodriguez, Joulin, Grave, and Lample}]{Touvron2023LLaMAOA}
Hugo Touvron, Thibaut Lavril, Gautier Izacard, Xavier Martinet, Marie-Anne Lachaux, Timoth{\'e}e Lacroix, Baptiste Rozi{\`e}re, Naman Goyal, Eric Hambro, Faisal Azhar, Aur'elien Rodriguez, Armand Joulin, Edouard Grave, and Guillaume Lample. 2023.
\newblock Llama: Open and efficient foundation language models.
\newblock \emph{ArXiv}, abs/2302.13971.

\bibitem[{Trinh and Le(2018)}]{DBLP:journals/corr/abs-1806-02847}
Trieu~H. Trinh and Quoc~V. Le. 2018.
\newblock \href {http://arxiv.org/abs/1806.02847} {A simple method for commonsense reasoning}.
\newblock \emph{CoRR}, abs/1806.02847.

\bibitem[{West et~al.(2022)West, Bhagavatula, Hessel, Hwang, Jiang, Le~Bras, Lu, Welleck, and Choi}]{west-etal-2022-symbolic}
Peter West, Chandra Bhagavatula, Jack Hessel, Jena Hwang, Liwei Jiang, Ronan Le~Bras, Ximing Lu, Sean Welleck, and Yejin Choi. 2022.
\newblock \href {https://doi.org/10.18653/v1/2022.naacl-main.341} {Symbolic knowledge distillation: from general language models to commonsense models}.
\newblock In \emph{Proceedings of the 2022 Conference of the North American Chapter of the Association for Computational Linguistics: Human Language Technologies}, pages 4602--4625, Seattle, United States. Association for Computational Linguistics.

\bibitem[{West et~al.(2021)West, Bhagavatula, Hessel, Hwang, Jiang, {Le Bras}, Lu, Welleck, and Choi}]{West2021SymbolicKD}
Peter West, Chandra Bhagavatula, Jack Hessel, Jena~D. Hwang, Liwei Jiang, Ronan {Le Bras}, Ximing Lu, Sean Welleck, and Yejin Choi. 2021.
\newblock Symbolic {K}nowledge {D}istillation: from general language models to commonsense models.
\newblock In \emph{North American Chapter of the Association for Computational Linguistics}.

\bibitem[{Xu et~al.(2023)Xu, Guo, Duan, and McAuley}]{Xu2023BaizeAO}
Canwen Xu, Daya Guo, Nan Duan, and Julian McAuley. 2023.
\newblock Baize: An open-source chat model with parameter-efficient tuning on self-chat data.
\newblock \emph{ArXiv}, abs/2304.01196.

\bibitem[{Zellers et~al.(2019)Zellers, Holtzman, Bisk, Farhadi, and Choi}]{Zellers2019HellaSwagCA}
Rowan Zellers, Ari Holtzman, Yonatan Bisk, Ali Farhadi, and Yejin Choi. 2019.
\newblock Hella{S}wag: Can a machine really finish your sentence?
\newblock In \emph{Annual Meeting of the Association for Computational Linguistics}.

\bibitem[{Zhang et~al.(2020{\natexlab{a}})Zhang, Khashabi, Song, and Roth}]{Zhang2020TransOMCSFL}
Hongming Zhang, Daniel Khashabi, Y.~Song, and D.~Roth. 2020{\natexlab{a}}.
\newblock {TransOMCS}: From linguistic graphs to commonsense knowledge.
\newblock In \emph{IJCAI}.

\bibitem[{Zhang et~al.(2020{\natexlab{b}})Zhang, Kishore, Wu, Weinberger, and Artzi}]{Zhang2020BERTScoreET}
Tianyi Zhang, V.~Kishore, Felix Wu, Kilian~Q. Weinberger, and Yoav Artzi. 2020{\natexlab{b}}.
\newblock Bertscore: Evaluating text generation with bert.
\newblock \emph{ArXiv}, abs/1904.09675.

\bibitem[{Zhou et~al.(2022)Zhou, Cho, Jandaghi, Lee, Lin, Pujara, and Ren}]{Zhou2022ReflectNR}
Pei Zhou, Hyundong~Justin Cho, Pegah Jandaghi, Dong-Ho Lee, Bill~Yuchen Lin, Jay Pujara, and Xiang Ren. 2022.
\newblock Reflect, not reflex: Inference-based common ground improves dialogue response quality.
\newblock In \emph{Conference on Empirical Methods in Natural Language Processing}.

\end{thebibliography}
 \bibliographystyle{acl_natbib}

\clearpage
\appendix

\section{Manual Cluster Analysis of Queries}
\label{sec:app-queries}

To understand the contents of \dataset, we analyze the top 100 surface form queries by total count in \dataset. We cluster these queries by hand into semantically related/equivalent groups, and then further take the top 10 of these groups, displayed in the main paper text. In Table~\ref{tab:query_clusters}, we include all queries in the the top 10 clusters along with with counts and total counts per cluster. 

\begin{table*}[t!]
\small
\centering 
\begin{tabular}{ l l}
\textbf{What time is it?} & 800\\
What time of day is it?	& 6153 \\
What is the time of day? & 458 \\
What time of the day is it? & 328 \\
\textit{Total} & 7739 \\
 & \\
\textbf{Who is PersonX?} & 2333 \\
What is an attribute of PersonX? & 434 \\
Who are PersonX and PersonY? & 257 \\
What is PersonX? & 233 \\
Who is PersonY? & 825 \\
\textit{Total} & 4082 \\
 & \\
\textbf{What is the weather like?} & 2960 \\
What's the weather like? & 382 \\
What is the weather like outside? & 351 \\
How is the weather? & 261 \\
\textit{Total} & 3954 \\
 & \\
\textbf{What is the prerequisite for this situation?} & 640 \\
What is a prerequisite for this situation? & 517 \\
What is a prerequisite for this event? & 345 \\
\textit{Total} & 1502 \\
 & \\
\textbf{What is the consequence of the situation?} & 314 \\ 
What's a potential consequence of this situation? & 311 \\
What is a potential consequence of this situation? & 225 \\
What is the consequence of this situation? & 197 \\
What is a consequence of this situation? & 190 \\
What could be a consequence of this situation? & 130 \\
\textit{Total} & 1367 \\
 & \\
\textbf{What is the counterfactual of the situation?} & 570 \\
What is the counterfactual of this situation? & 202 \\
What is a counterfactual of the situation? & 163 \\
What is a counterfactual of this situation? & 125 \\
\textit{Total} & 1060 \\
 & \\
\textbf{What will happen next?} & 268 \\
 What will the person do next? & 211 \\
 What will PersonX do next? & 210 \\
What will they do next? & 156 \\
What might happen next? & 155 \\
\textit{Total} & 1000 \\
 & \\
\textbf{What is the occasion?} & 712 \\
 What's the occasion? & 149 \\
\textit{Total} & 861 \\
 & \\
\textbf{What is the relationship between PersonX and PersonY?} & 655 \\
What is their relationship? & 198 \\
\textit{Total} & 853 \\
 & \\
\textbf{Where are they?} & 193 \\
Where is PersonX? & 233 \\
What is the setting? & 223 \\
Where is this taking place? & 135 \\
\textit{Total} & 784 \\
\end{tabular}
    \caption{The surface forms and counts included in the top 10 clusters of analyzed queries.}
\label{tab:query_clusters}
\end{table*}

\subsection{Automatic Evaluation of Generation}

\begin{table*}[t!]
    \centering\footnotesize
     \scalebox{.98}{
    \begin{tabular}{l|cccc|cccc}
\toprule
 & \multicolumn{4}{c|}{\textbf{BLEU}} & \multicolumn{4}{c}{\textbf{BERTScore}} \\
                  system &  $\alpha$NLG & Reflect & TellMeWhy & \atomictwenty & $\alpha$NLG & Reflect & TellMeWhy & \atomictwenty \\
\midrule

LLaMA-7B &  0.8 & 0.4 & 2.1 & 0.1 & 85.2 & 83.2 & 85.9 &81.7\\
LLaMA-13B & 1.0 & 0.5 & 4.6 & 0.1 & 85.4 & 83.8 & 86.5 & 81.7\\
T0 & 1.3 & 3.2 & 9.0 & 0.5& 87.2 & 88.7 & 89.1 & 85.3\\
Alpaca-7b & 1.3 & 1.1 & 6.4 & 0.3 & 88.7 & 87.8 & 89.4 & 83.5\\
Alpaca-13B & 2.4 & 1.2 & 6.9 & 0.2 & 88.8 & 87.4 & 89.2 & 83.3\\
Flan-Ul2 & 4.3 & 3.4 & 5.7 & 0.5 & 90.0 & 86.5 & 85.9 & 85.3\\
Flan-T5$_{xxl}$ & 4.3 & 4.4 & 10.8 & 0.5 & 90.0 & 88.2 & 90.2 & 86.3\\
\midrule
\modelgen & 3.4 & 3.7 & 10.8 & 0.6& 89.7 & 88.5 & 90.8 & 85.8\\
\bottomrule
    \end{tabular}}
    \caption{Comparison of baselines and the \modelgen using automatic scores BLEU \cite{papineni2002bleu} and BERTScore \cite{Zhang2020BERTScoreET}. Automatic metrics do not seem to agree with human evaluation, and show less clear variation. 
    }
    \label{tbl:generation_auto}
\end{table*}

We also include automatic evaluation with 2 metrics in Table~\ref{tbl:generation_auto}. We find these values show a much less distinct spread, with no model taking a clear lead over others. The seemingly lower information and general unreliability of automatic metrics was a motivation in mainly considering human evaluation.

\section{Data Generation}
\subsection{Context Generation Prompts}
\label{sec:app-context}

Below are the 21 prompts used for doing context generation (delimited with ''')

\begin{lstlisting}
Generate 20 events.

1. Event:

'''
Generate 20 common events.

1. Event:
'''
Generate 20 everyday events.

1. Event:
'''
Generate 20 events that happen often.

1. Event:
'''
Generate 20 events that happen sometimes.

1. Event:
'''
Generate 20 events that include a person or people.

1. Event:
'''
Generate 20 everyday events about PersonX (one per line). It may also involve other entities, such as PersonY.

1. Event:
'''
Generate 20 situations.

1. Situation:
'''
Generate 20 common situations.

1. Situation:
'''
Generate 20 everyday situations.

1. Situation:
'''
Generate 20 situations that happen often.

1. Situation:
'''
Generate 20 situations that happen sometimes.

1. Situation:
'''
Generate 20 situations that include a person or people.

1. Situation:
'''
Generate 20 everyday situations about PersonX (one per line). It may also involve other entities, such as PersonY.

1. Situation:
'''
Generate 20 situations. They should be complex and include multiple parts. (One per line)

1. Situation:
'''
Generate 20 common situations. They should be complex and include multiple parts. (One per line)

1. Situation:
'''
Generate 20 everyday situations. They should be complex and include multiple parts. (One per line)

1. Situation:
'''
Generate 20 situations that happen often. They should be complex and include multiple parts. (One per line)

1. Situation:
'''
Generate 20 situations that happen sometimes. They should be complex and include multiple parts. (One per line)

1. Situation:
'''
Generate 20 situations that include a person or people. They should be complex and include multiple parts. (One per line)

1. Situation:
'''
Generate 20 everyday situations about PersonX (one per line). It may also involve other entities, such as PersonY. They should be complex and include multiple parts. (One per line)

1. Situation:
\end{lstlisting}

\subsection{Relation Generation Prompts}
Below are the prompts for generating relations. To promote diversity, the number of examples were randomly selected from Uniform(1,10) and were shuffled. Some contexts come from ROCStories \cite{mostafazadeh-etal-2016-corpus} and \cite{west-etal-2022-symbolic}, while others are handwritten. All questions and inferences are hand-written.
When prompting `gpt-3.5-turbo`, we provide the instructions "Given a situation... answer" as the system message, the Context as a user message, and the ten generated queries/inferences as the system response.

\label{sec:app-inference}
\subsection{With queries}
\begin{lstlisting}
Given a situation, ask and answer ten (10) relevant questions that require commonsense or a world model. Some examples may include potential consequences, explanations, prerequisites or reactions, attributes, or counterfactuals. The commonsense facts may be about actors, actions, events, or ideas in the passage. The examples should be high-quality and things that are true. Please give a plausible answer at all times instead of just saying that it depends. Only ask questions that will have a relevant, commonsense answer.


Alisa and her family lived in Florida. They heard a hurricane was coming. They decided to evacuate to a relative's house. They arrived and learned from the news that it was a terrible storm.
1. What will happen now? They will wait out the storm.
2. How does Alisa feel? She is probably relieved to be out of the hurricane's path.
3. What would have happened if Alisa and her family had not evacuated? They would have been in the storm.
4. Why did they decide to evacuate to a relative's house? They wanted to be in a safe place.
5. Alisa's family is what? Responsible
6. What might have prevented them from fleeing? If they had not heard about the hurricane, or if they had no way to get to a relative's house.
7. They would not have fled if they were not what? Cautious
8. Where does their relative live? Somewhere safe from the hurricane.
9. Should they have fled even if the storm hadn't been bad? Yes, because they might have not been able to leave if the hurricane got worse.
10. How could you describe their relative? Kind

A robber steals from a bank.
1. What are some potential characters in the situation? Robber, bank teller/worker, customers
2. Tell me something about the robber? The robber is probably armed
3. What does the robber have? The robber probably has a getaway car
4. What does the bank teller feel? The bank teller is probably scared
5. What might happen to the robber? The robber could go to jail
6. What does the bank have? The bank might have a security system
7. Before this, did the robber do anything? The robber probably planned this in advance
8. As a consequence, what will happen? After, the robber will have the money
9. What happens before this? The robber tells the bank teller to give them the money
10. How much money does the robber get? A lot of money

The woman enters the elevator
1. What did the woman have to do before? The woman had to press the button for the elevator to come to her floor
2. What is the woman's goal? The woman wants to go to a different floor
3. What will the woman do next? The woman will press the button for the floor she wants to go to
4. What could hinder this situation? The woman wants to take the stairs to be healthy
5. Is she alone? She may or may not be alone, since there could be other people in the elevator.
6. What does the woman see in the elevator? Buttons to different floors
7. What does the woman feel? The woman could feel impatient at having to wait for an elevator
8. As a consequence, what will happen? The woman will arrive at her desired floor
9. What could prevent this from happening? The elevator is out of service
10. Where are elevators located? Multi-story buildings

Emma has a big exam tomorrow. She got so stressed, she pulled an all-nighter. She went into class the next day, weary as can be. Her teacher stated that the test is postponed for next week.
1. How does Emma feel about this? Emma is probably relieved
2. Why might Emma be frustrated? Emma could be frustrated because she stayed up all night studying for nothing
3. What is the consequence of the situation? Emma will have more time to study
4. What is the prerequisite for this situation? Emma needed to have a big exam
5. Tell me what Emma will do next. Emma will probably go home and sleep.
6. What did Emma do before this? Emma was studying for her exam
7. Why did the teacher postpone the exam? The teacher may have postponed the exam because not everyone was ready.
8. What is an attribute of Emma? Emma is a procrastinator.
9. What is an attribute of Emma's teacher? flexible
10. What is the counterfactual of the situation? If Emma didn't have a big exam, she wouldn't have pulled an all-nighter.

Karen was assigned a roommate her first year of college. Her roommate asked her to go to a nearby city for a concert. Karen agreed happily. The show was absolutely exhilarating.
1. What's something we can infer about Karen? Karen likes music
2. What will happen because of this? Karen and her roommate will be better friends
3. How might this have been prevented? If Karen's roommate was shy, she might not have asked Karen to go to a concert
4. How old is Karen? Young adult
5. Why did Karen agree happily? Karen wanted to get to know her roommate better, make friends, and enjoy a concert
6. How did Karen and her roommate get to the concert? By car or public transportation
7. What's a potential consequence of this situation? Karen might have fun and meet new people
8. How does Karen feel? Karen is pleased
9. What does Karen's roommate think of her? The roommate thinks Karen is cool
10. When is the concert? The concert is likely at night

Ivette misplaced her phone at her grandparents.
1. What did Ivette do before this? Ivette was at her grandparents
2. How does Ivette feel? Ivette feels frustrated
3. What will Ivette do next? Ivette will look for her phone
4. What could hinder this situation? If Ivette was more careful
5. Ivette is what? Young
6. What would make this situation harder for Ivette? Her phone is turned off.
7. Where might her phone be? Ivette's phone could be in the house, outside, or in the car.
8. Did Ivette mean to lose her phone? No
9. What is a consequence of the situation? Ivette will have to buy a new phone
10. What would remedy the situation? Finding Ivette's phone

PersonX takes PersonY back to the hospital
1. Why did PersonX take PersonY back to the hospital? PersonY was not feeling well
2. What happened before this? PersonY was discharged from the hospital
3. What is PersonX and PersonY's relationship to eachother? They are either friends or family.
4. What would make this hard? PersonX doesn't have a car.
5. Next, what will happen? PersonY will receive medical care.
6. What happened before? PersonY asked PersonX to take them to the hospital.
7. What is PersonX? PersonX is kind
8. What is PersonY? PersonY is sick
9. Where are they? They are in a car
10. What is a result? PersonY will get better

Mila and her family lived in Florida. They heard a hurricane was coming. They decided to evacuate to a relative's house. They arrived and learned from the news that it was a terrible storm.
1. What will happen now? They will wait out the storm.
2. How does Mila feel? She is probably relieved to be out of the hurricane's path.
3. What would have happened if Mila and her family had not evacuated? They would have been in the storm.
4. Why did they decide to evacuate to a relative's house? They wanted to be in a safe place.
5. Mila's family is what? Responsible
6. What might have prevented them from fleeing? If they had not heard about the hurricane, or if they had no way to get to a relative's house.
7. They would not have fled if they were not what? Cautious
8. Where does their relative live? Somewhere safe from the hurricane.
9. Should they have fled even if the storm hadn't been bad? Yes, because they might have not been able to leave if the hurricane got worse.
10. How could you describe their relative? Kind

Alegra coyly smiled at the boy as he walked in.
1. Why did Alegra smile at the boy? Alegra was interested in him.
2. What will the boy do? The boy will notice Alegra.
3. What is Alegra's relationship to the boy? They are strangers.
4. What will happen if Alegra keeps smiling at the boy? The boy might talk to her.
5. If the boy doesn't talk to her, how will Alegra feel? Alegra will feel awkward.
6. What is the difference between a coy smile and a regular smile? A coy smile is more flirtatious.
7. How could Alegra be described? Confident
8. Where is this probably located? In a public place
9. Alegra is probably what age? A teenager or young adult
10. What would prevent this from happening? Alegra is scared to put herself out there

PersonX crosses the road
1. What is PersonX? A pedestrian
2. What could prevent this from happening? This could be prevented if there was no crosswalk.
3. What is a prerequisite for this event? A prerequisite for this event is that PersonX wants to cross the road.
4. What is something that could happen? PersonX gets hit by a car
5. If this didn't happen, what would happen? If this didn't happen, PersonX would not get to where they need to go.
6. What actors might be in this situation? PersonX, drivers, other pedestrians
7. What might PersonX be thinking? PersonX might be thinking that they need to get to the other side of the road.
8. What could be true to make PersonX reckless? PersonX crosses when there are lots of cars and no crosswalk
9. What could be true to make PersonX cautious? PersonX waits carefully for the walk signal and looks both ways.
10. What do people do before crossing the road? People might look both ways to check for cars.

Honor decides whether to bike or walk to school.
1. In what situation would Honor choose to walk to school? It is raining outside.
2. What is Honor? a student
3. How could Honor be described? Unsure
4. What would make this improbable? Honor lives very far away from the school.
5. Why might Honor choose to bike over walk? It is faster.
6. What will happen if Honor can't make up his mind? Honor will be late for school.
7. What is a possible reason for why Honor can't decide? He is feeling lazier today.
8. Either way, Honor will what? Get exercise
9. What is the difference between biking and walking? Biking is faster but requires more effort.
10. What is the weather? It might be sunny.
\end{lstlisting}

\subsection{Without queries}
\begin{lstlisting}
List ten (10) commonsense facts about each situation. Some examples may include potential consequences, explanations, prerequisites or reactions. The commonsense facts may be about actors, actions, events, or ideas in the passage. The outputs could also include counterfactuals or things that could hinder the event from happening. The examples should be high-quality and things that are true.

PersonX crosses the road
1. PersonX is probably going to the other side
2. Cars are on the road
3. Before this can happen, PersonX looks both ways to make sure it's safe
4. PersonX probably has a destination
5. PersonX is probably walking
6. This wouldn't happen if there wasn't a crosswalk
7. After, PersonX will be on the other side
8. If PersonX is jaywalking, they might get hit by a car
9. PersonX might use a crosswalk signal
10. This couldn't happen if the person wasn't near a road

A robber steals from a bank.
1. The robber is probably armed
2. The robber probably has a getaway car
3. The bank teller is probably scared
4. This is illegal
5. The robber could go to jail
6. The bank might have a security system
7. The robber probably planned this in advance
8. After, the robber will have the money
9. Before this happens, the robber tells the bank teller to give them the money
10. The robber might wear a mask

Addilyn and her family lived in Florida. They heard a hurricane was coming. They decided to evacuate to a relative's house. They arrived and learned from the news that it was a terrible storm.
1. They may have left valuables behind
2. They may come back to a destroyed house
3. They were smart to evacuate
4. If they didn't evacuate, they might have died
5. The hurricane was very bad
6. Now, they will wait out the storm
7. They went to their relatives house because they wanted to be in a safe place
8. They wouldn't have fled if they had not heard about the hurricane
9. The relative lives somewhere safe from the hurricane
10. Their relative is kind for letting them stay over

Fatima was assigned a roommate her first year of college. Her roommate asked her to go to a nearby city for a concert. Fatima agreed happily. The show was absolutely exhilarating.
1. Fatima has a roommate
2. Fatima likes music
3. As a result, Fatima and her roommate will be better friends
4. Fatima enjoyed the concert
5. In the future, Fatima may want to go to more concerts
6. Fatima may be more likely to spend time with her roommate
7. Fatima's roommate is considerate
8. Fatima's roommate is probably also a student
9. The roommate thinks that Fatima is cool
10. They got to the concert using a car or public transportation

Loretta misplaced her phone at her grandparents.
1. As a result, Loretta may be stressed.
2. Loretta may have to buy a new phone.
3. This event may have ruined Loretta's weekend.
4. This wouldn't happen if Loretta was more careful.
5.. Now, Loretta will probably look for her phone.
6. It will be expensive to replace her phone if it is lost.
7. Loretta is young.
8. This situation would be worse if Loretta's phone was turned off.
9. Things could be better if Loretta finds her phone
10. Loretta did not mean to lose her phone

SAN FRANCISCO - Charlotte's husband, Maxwell, was violently assaulted by a man who broke into the couple's home in San Francisco early Friday morning, the police said. The authorities identified the suspect as Lozen, 42, and said they were investigating a possible motive.
1. Lozen could be mentally ill
2. Maxwell was likely asleep when the attack happened
3. Lozen is either in custody or being searched for by the police
4. The breaking and entering was likely planned
5. Charlotte was probably not attacked
6. This would have been a frightening experience for Charlotte and Maxwell
7. If Lozen is caught, he will likely go to jail
8. Lozen's motive might have been personal
9. This wouldn't have happened if Lozen were not violent
10. Home invasions are usually premeditated

The woman enters the elevator
1. Before, the woman pushed the button for the elevator
2. The woman is going to a different floor
3. After, the woman will push the button for her floor
4. Then, she will press the button for her desired floor
5. First, the woman will wait for other people to walk out of the elevator
6. The woman might have been impatient if she had to wait for a long time
7. This couldn't happen if the elevator were out of service
8. The woman would not have done this if she wanted to take the stairs to be healthy
9. The woman may have been in a hurry
10. She is in a multi-story building

PersonX takes PersonY back to the hospital.
1. PersonY has been to the hospital before
2. The goal of PersonX was to help PersonY
3. Before this can happen, PersonY must ask PersonX to take them to the hospital
4. PersonY hopes to get medical care
5. PersonY may have been injured before this
6. This couldn't happen if PersonX does not have a car
7. PersonY is sick in some way
8. Going to the hospital may be expensive
9. This probably wouldn't happen if PersonY wasn't sick
10. PersonX cares about PersonY

Michaela has a big exam tomorrow. She got so stressed, she pulled an all-nighter. She went into class the next day, weary as can be. Her teacher stated that the test is postponed for next week.
1. Michaela is relieved that she doesn't have to take the test today
2. Michaela is sad because she worked hard to prepare and in the end didn't have to
3. When Michaela stayed up all night she was studying
4. The teacher probably postponed the exam because not everyone was ready.
5. Next week, Michaela will have to study again
6. Michaela may do better on the exam next week because she will have more time to prepare
7. If the exam was today, Michaela would have done poorly
8. The test is in a subject that Michaela is struggling in
9. Michaela is a procrastinator
10. If Michaela hadn't stayed up all night, she would not be tired

Keanu decides whether to bike or walk to school.
1. Keanu might not choose to bike if it's raining outside
2. Keanu is a student
3. They are unsure
4. If Keanu doesn't make up their mind, Keanu will be late for school
5. This is because Keanu is feeling lazy today
6. Either way, Keanu will get exercise
7. Biking is faster than walking but requires more effort
8. This wouldn't happen if Keanu were more decisive
9. This would be hard if Keanu lived very far away from the school
10. Keanu might choose to bike if it's a nice day outside
    
\end{lstlisting}

\section{MTurk Templates}
\label{sec:app-amt}

\begin{figure*}[!b]
    \centering
    \makebox[\textwidth]{\includegraphics[height=0.7\textheight]{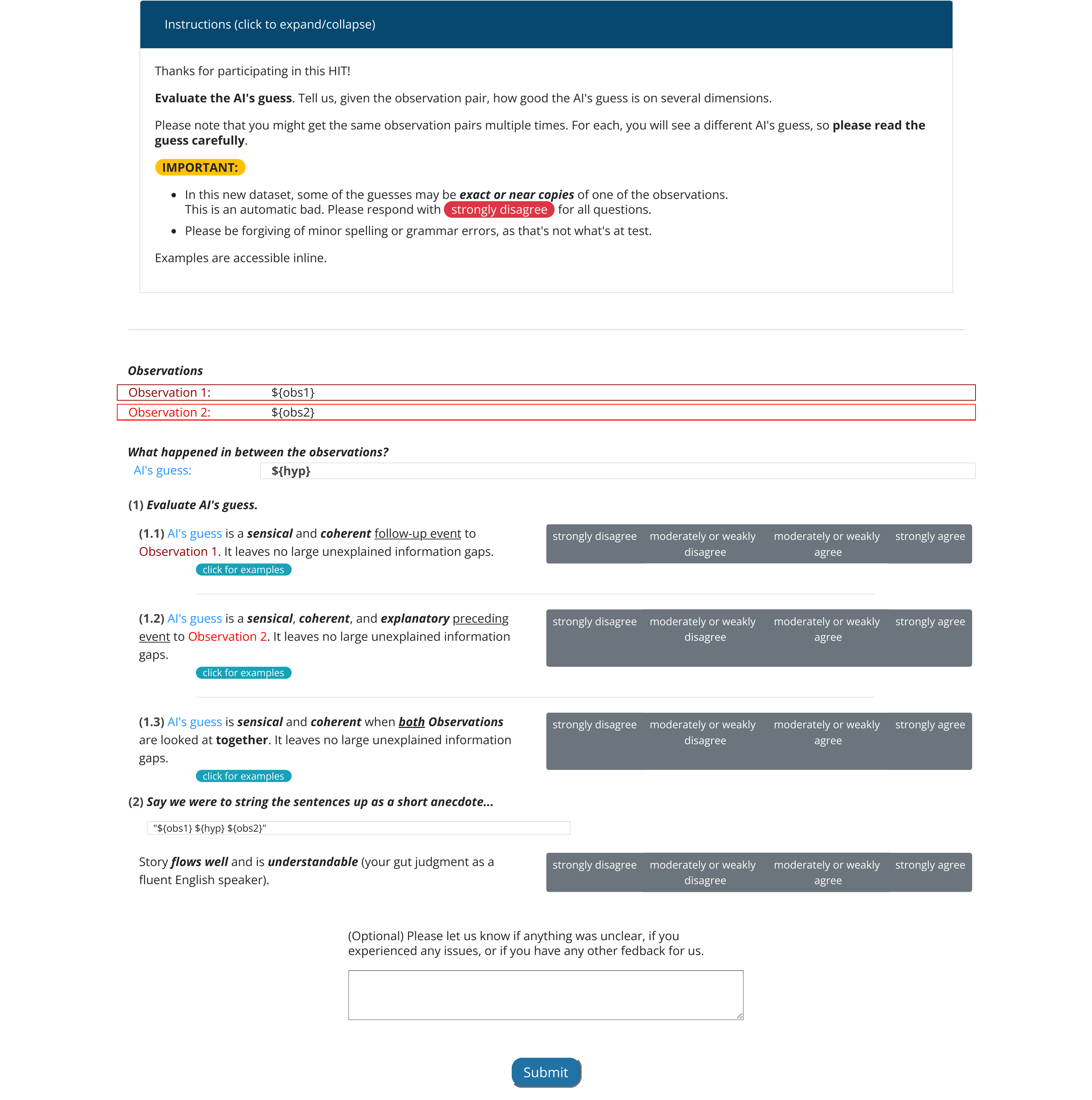}}
    \caption{MTurk template for $\alpha$ NLG.}
    \label{fig:anlg-template}
\end{figure*}

\begin{figure*}[!b]
    \centering
    \makebox[\textwidth]{\includegraphics[height=0.8\textheight]{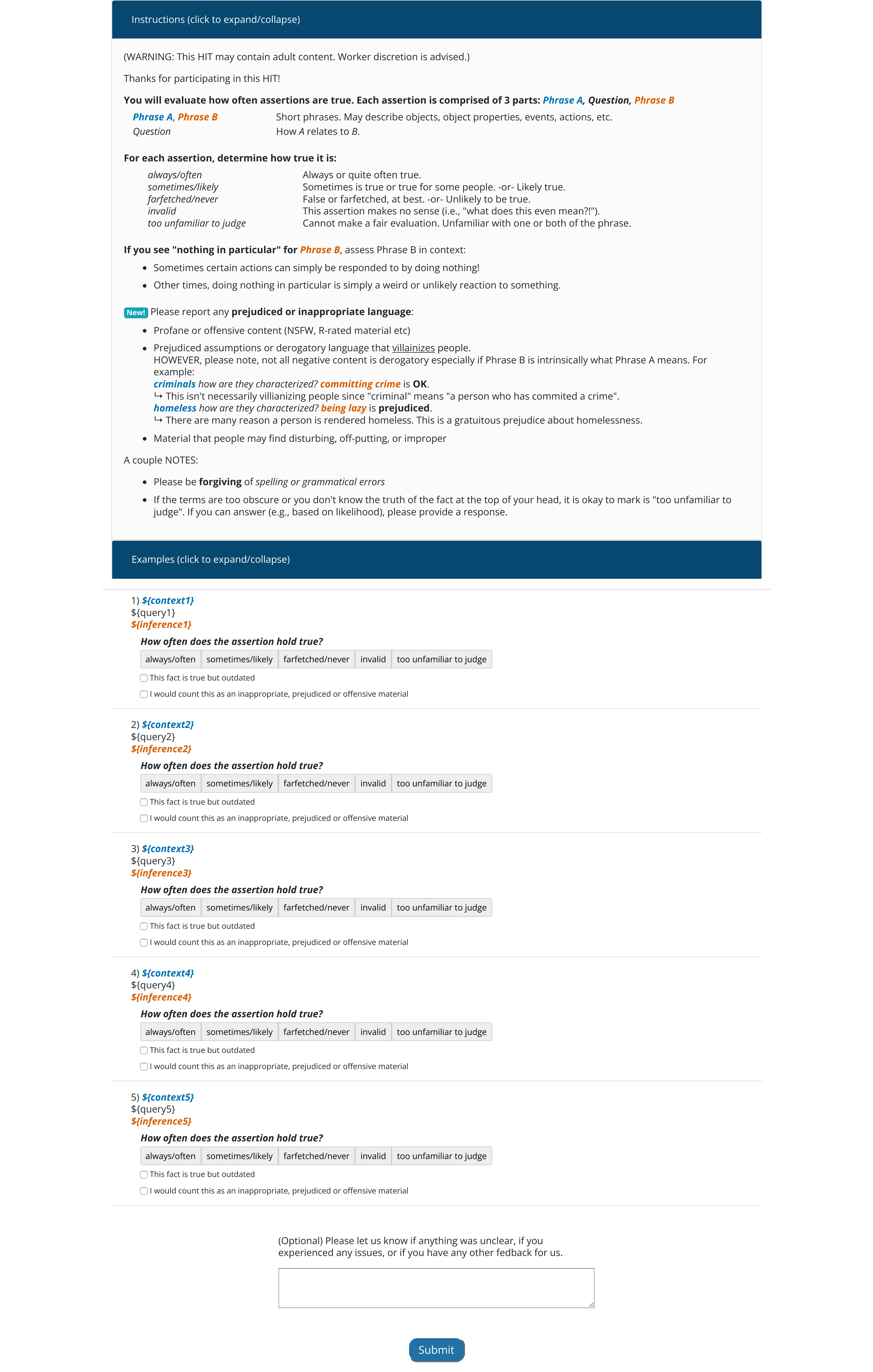}}
    \caption{MTUrk template for CQI.}
    \label{fig:anlg-template2}  
\end{figure*}



\end{document}